\documentclass[10pt,twocolumn,letterpaper]{article}

\usepackage[pagenumbers]{cvpr} % To force page numbers, e.g. for an arXiv version

\definecolor{cvprblue}{rgb}{0.21,0.49,0.74}
\usepackage[pagebackref,breaklinks,colorlinks,allcolors=cvprblue]{hyperref}

\usepackage{enumitem}
\usepackage{booktabs}
\usepackage{tabularx}
\usepackage{ragged2e}
\usepackage[accsupp]{axessibility}  % Improves PDF readability for those with disabilities.

\def\paperID{} % *** Enter the Paper ID here
\def\confName{CVPR}
\def\confYear{2026}

\title{Beyond Pixels: Benchmarking and Reward-Based Assessing Framework\\
for Visual Spatial Aesthetics}

\author{
Yuan Gao$^{1}$, Jin Song$^{1}$, Yiyun Fei$^{1}$, Gongzhe Li$^{2}$, Ruigao Yang$^{1\dagger}$\\
$^{1}$Alibaba Group  $^{2}$The Chinese University of Hong Kong, Shenzhen \\
{\tt\small gaoyuan20@mails.ucas.ac.cn, songjin.song@alibaba-inc.com, yunhun.fyy@alibaba-inc.com} \\
{\tt\small gongzheli1@link.cuhk.edu.cn, ruigao.yrg@alibaba-inc.com}
}

\begin{document}
\maketitle

% 通讯作者说明
\def\thefootnote{$\dagger$}\footnotetext{Corresponding author.}
\def\thefootnote{\arabic{footnote}}

\begin{abstract}
% The ABSTRACT is to be in fully justified italicized text, at the top of the left-hand column, below the author and affiliation information.
% Use the word ``Abstract'' as the title, in 12-point Times, boldface type, centered relative to the column, initially capitalized.
% The abstract is to be in 10-point, single-spaced type.
% Leave two blank lines after the Abstract, then begin the main text.
% Look at previous \confName abstracts to get a feel for style and length.

In recent years, Image Quality Assessment (IQA) for AI-generated images (AIGI) has advanced rapidly; however, existing methods primarily target portraits and artistic images, lacking a systematic evaluation of interior scenes. We introduce Spatial Aesthetics, a paradigm that assesses the aesthetic quality of interior images along four dimensions: layout, harmony, lighting, and distortion. We construct SA-BENCH, the first benchmark for spatial aesthetics, comprising 18,000 images and 50,000 precise annotations. Employing SA-BENCH, we systematically evaluate current IQA methodologies and develop SA-IQA, through MLLM fine-tuning and a multidimensional fusion approach, as a comprehensive reward framework for assessing spatial aesthetics. We apply SA-IQA to two downstream tasks: (1) serving as a reward signal integrated with GRPO reinforcement learning to optimize the AIGC generation pipeline, and (2) Best-of-N selection to filter high-quality images and improve generation quality. Experiments indicate that SA-IQA significantly outperforms existing methods on SA-BENCH, setting a new standard for spatial aesthetics evaluation. Code is available at \url{https://github.com/AlibabaResearch/SA-IQA}.

\end{abstract}    
\section{Introduction}
\label{sec:intro}

\begin{figure*}[ht]
    \centering
    \includegraphics[width=\textwidth]{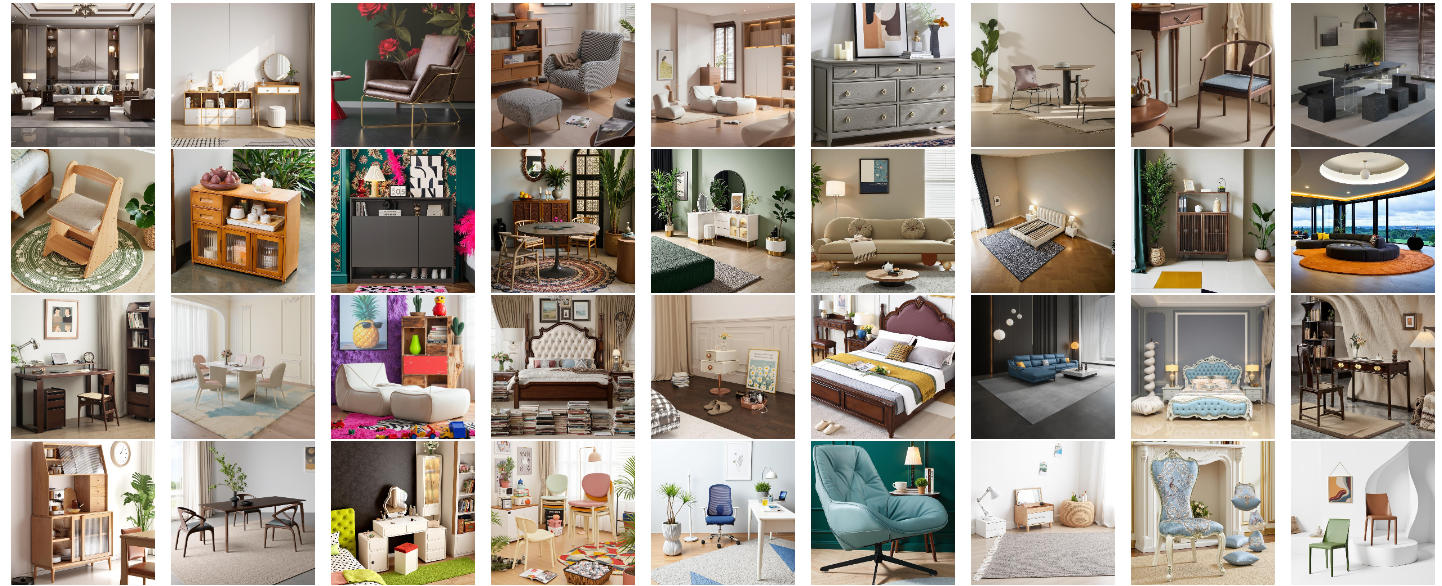}
    \caption{\textbf{Visualizing the SA-BENCH dataset.} 
From top to bottom, each row corresponds to the results of different models in the dataset: the first row uses SD1.5-Inpaint \cite{sd}, the second row is SDXL-BrushNet \cite{brushnet}, the third row is FLUX-Inpaint \cite{labs2025flux}, and the fourth row is FLUX EasyControl \cite{zhang2025easycontrol}.}
    \label{fig:first}
\end{figure*}

The growth of Generative Artificial Intelligence (AIGC) \cite{sd,sd3,sdxl} has made Image Quality Assessment (IQA) \cite{ssim} increasingly important \cite{wu2023assessor360,ying2020patches,chow2016review}. IQA has two main roles: one is to filter low-quality generated content, and the second is to be used as a human preference alignment signal for post-training generative models \cite{wu2023better}. Therefore, the evaluation content of IQA has also evolved from early traditional image quality (such as blur or noise) \cite{yang2019survey} to the current stage's complex human preferences, such as image aesthetics \cite{Deng2022Rethinking}, human anatomy \cite{AGHI-QA}, text-image alignment \cite{Yu2024SFIQA}, and the latest instruction following in image editing \cite{gong2025onereward}.

Along with the change in IQA's research focus, the research methods have also evolved from being based on early pre-trained models like CLIP \cite{wang2023clipiqa} to leveraging Multimodal Large Language Models (MLLMs), like Q-Align \cite{qalign}, which perform finer-grained, human-aligned evaluations by utilizing the capabilities of large pre-trained models.

AI is being rapidly and widely adopted in applications like interior design and furniture e-commerce. However, existing IQA methods are often trained on general-purpose datasets \cite{kirstain2023pickapicopendatasetuser, li2024aigiqa} that cover mass aesthetic preferences. While specialized benchmarks for domains like human figures \cite{AGHI-QA} or artistic style \cite{yi2023towards} exist, there is currently no dataset or corresponding IQA method dedicated to interior design spatial aesthetics. 

Evaluating ``Spatial Aesthetics'' is inherently multi-factorial. We formulate it as a practical framework for AI-generated interior scenes along four dimensions---layout, harmony, lighting, and distortion---grounded in interior design practice and perceptual realism. While not exhaustive, this decomposition is interpretable and suitable for expert annotation and reward modeling.

To comprehensively address the application of IQA in spatial aesthetics, evaluate and improve the quality of AI-generated interior images, we first construct a high-quality dataset with multi-dimensional human annotations and train a IQA model based on this dataset. Specifically, our contributions are:

\begin{itemize}
    \item \textbf{SA-BENCH:} We define the Spatial Aesthetics assessment paradigm for AI-generated residential interior furnishing scenes along four dimensions---layout, harmony, lighting, and distortion---and construct the first benchmark comprising 18,000 interior images with 50,000 precise human annotations. The visualization of a portion of the dataset is shown in Figure \ref{fig:first}.

    \item \textbf{SA-IQA:} We introduce a Spatial Aesthetics IQA model that outputs calibrated, multi-dimensional rewards (layout, harmony, lighting, distortion). Built via MLLM fine-tuning with expert-aware instructions and a multidimensional fusion optimization, SA-IQA attains state-of-the-art PLCC/SROCC on SA-BENCH.

    \item \textbf{Downstream Applications:} We fully validate the effectiveness of our SA-IQA model in two representative downstream tasks: first, we integrate SA-IQA as a reward signal, in conjunction with GRPO, to optimize a prompt expansion module for image generation task, and second, in Best-of-N selection to filter low-quality images, significantly improving generation quality.
\end{itemize}

\section{Related Work}
\label{sec:related_work}

\begin{figure*}[!t]
    \centering
    % 使用您上传的图片文件名作为路径
    \includegraphics[width=0.95\textwidth]{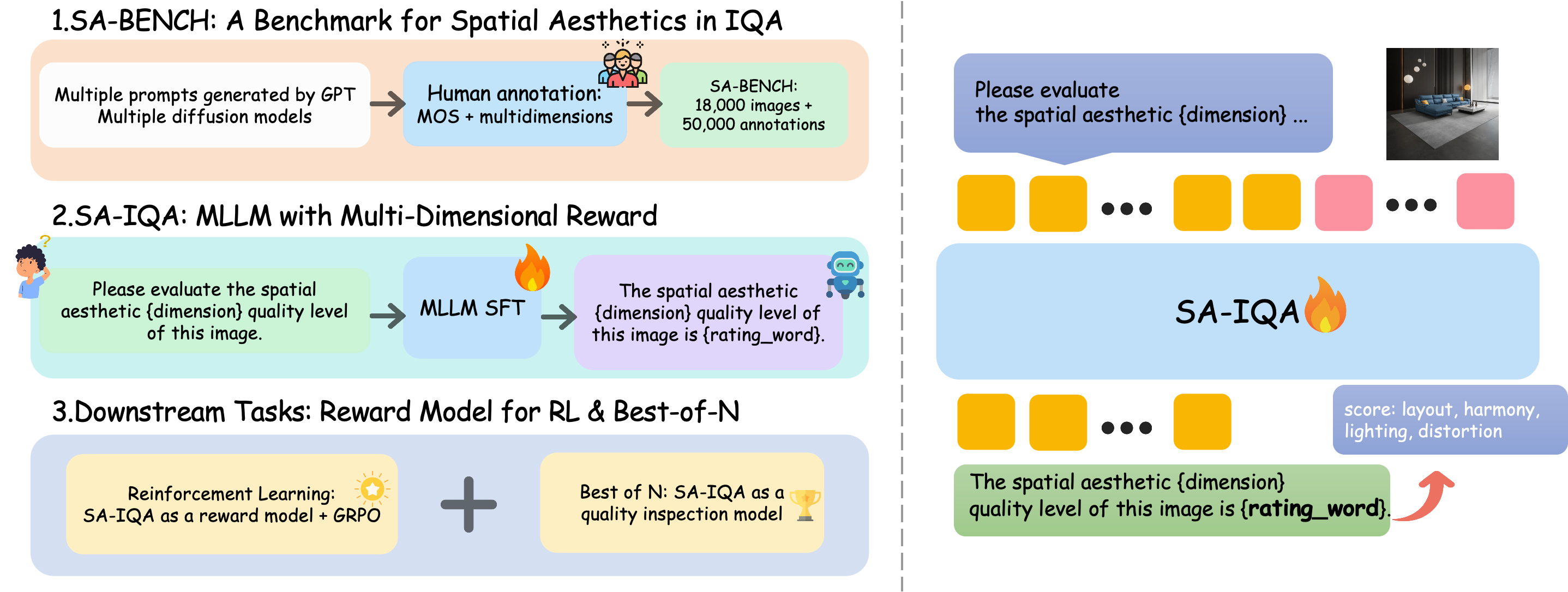}
    \caption{\textbf{Overview of the SA-IQA Framework.} The left panel depicts our three-stage workflow: establishing Spatial Aesthetics dimensions and constructing SA-BENCH, developing the SA-IQA reward model via MLLM fine-tuning, and deploying it for GRPO-based prompt optimization and Best-of-N selection. The right panel details the SA-IQA model's principle, showing how it processes an image and a dimension-conditioned query to predict multi-dimensional MOS scores, which are then calibrated and fused into a single spatial aesthetic score.}
    \label{fig:method_overview}
\end{figure*}

\subsection{Text-to-Image Generation}
The field of text-to-image (T2I) generation has grown rapidly. Open-source models like Stable Diffusion (SD) \cite{sd}, SDXL \cite{sdxl}, and SD3 \cite{sd3} continue to improve image realism. Commercial models such as Midjourney and Seedream4.0 \cite{seedream4.0} can already generate highly realistic images. The training pipeline for these models often includes pre-training, fine-tuning, and reinforcement learning (RL) stages to align with human preferences, which requires collecting large-scale preference data.

Although some open-source aesthetic preference datasets exist, such as ImageReward \cite{ImageReward} and HPSV2 \cite{HPSV2}, this data often reflects general preferences. This can lead to a lack of precision when working in specific domains, like interior design. To fix this, some methods collect domain-specific preference data. For instance, flux-krea \cite{flux1kreadev2025} collected its data in a ``very opinionated manner'' to match a specific aesthetic taste and a ``clear art direction.'' Similarly, other methods have used preference data focused on anatomical distortions to improve how models generate human anatomy \cite{AGHI-QA}.

\subsection{Image Quality Assessment}
Early AIGI-oriented IQA benchmarks have begun to formalize evaluation factors beyond a single quality score. 
AGIQA-3K~\cite{agiqa3k} and AIGCIQA2023~\cite{aigciqa_2023} adopt multidimensional annotations (e.g., perceptual quality, authenticity, and text-image correspondence), and AIGIQA-20K~\cite{li2024aigiqa} further scales this direction. 
However, these benchmarks are largely general-purpose and rarely disentangle domain-specific, spatially structured factors (e.g., layout coherence) that are crucial for interior scenes.
Recent efforts therefore move towards domain-specific IQA, exemplified by AGHI-QA~\cite{AGHI-QA}, which introduces detailed structural annotations for distorted human parts.

Meanwhile, MLLMs are increasingly used as IQA assessors to improve interpretability and alignment with human judgments. 
Q-Align~\cite{qalign} formulates quality prediction with discrete, text-defined levels, while Q-Insight~\cite{qinsight} leverages reinforcement learning to enhance robustness under limited supervision. 
GROUNDING-IQA~\cite{chen2024grounding} further extends IQA from global scoring to spatially grounded assessment by coupling quality descriptions with localized regions.

Different from prior general-purpose AIGI-IQA benchmarks and distortion-centric evaluators, we focus on interior images where aesthetic quality strongly depends on spatial organization.
We thus target Spatial Aesthetics along layout, harmony, lighting, and distortion, and build an MLLM-based evaluator on a dedicated benchmark to enable multidimensional, spatially-aware assessment for generation-time optimization.
\section{Method}
\label{sec:method}

We propose SA-IQA, a novel framework designed to address the unique challenges of interior spatial aesthetics evaluation. As shown in Figure~\ref{fig:method_overview}, it comprises three stages: (1) the creation of ``SA-BENCH'', a specialized multi-dimensional benchmark through rigorous human annotation; (2) the development of the ``SA-IQA'' model via MLLM Supervised Fine-Tuning (SFT) and a multi-dimensional fusion module; and (3) the application of SA-IQA in ``Downstream Tasks'' (Reinforcement Learning optimization and Best-of-N selection) to enhance the quality of AIGC outputs.

\subsection{SA-BENCH}
\subsubsection{Benchmark overview}
We introduce SA-BENCH, a benchmark tailored to spatial aesthetics of interior AIGI with four dimensions—layout, harmony, lighting, and distortion—and high-quality Mean Opinion Score (MOS) annotations. SA-BENCH focuses on residential interior furnishing scenes. It comprises 17753 images and 50476 precise annotations, establishing the first large-scale, multi-dimensional benchmark for spatial-aesthetics IQA in this domain.

\begin{table*}[t]
\centering
\caption{Comparison of different AIGC Quality Datasets.}
\label{tab:agi_quality_databases}
\begin{tabular}{@{}llllrrl@{}}
\toprule
\textbf{Database} & \textbf{Annotation Type} & \textbf{Dimensions} & \textbf{Images} & \textbf{Annotations} & \textbf{Generation Models} \\
\midrule
DiffosonDB\cite{wang2023diffusiondb} & No & No & 1,819,808 & 0 & 1 \\
AGIQA-1K \cite{agiqa-1k} & MOS & Perception & 1,080 & 23,760 & 2 \\
Pick-A-Pic \cite{kirstain2023pickapicopendatasetuser} & Preference & Overall & 500,000 & 500,000 & 3 \\
HPS\cite{HPS} & Preference & Overall & 98,807 & 98,807 & 1 \\
ImageReward \cite{imagereward:2023} & Ranking & Perception; Alignment & 136,892 & 410,676 & 7 \\
AGIQA-3K \cite{agiqa3k} & MOS & Perception; Alignment & 2,982 & 125,244 & 8 \\
SA-BENCH & MOS & Spatial Aesthetics & 17,753 & 50,476 & 4 \\
\bottomrule
\end{tabular}
\end{table*}

\subsubsection{Data Construction}
We begin with a large pool of real-world interior photographs as raw data. To reduce redundancy, we first extract image features using DreamSim \cite{fu2023dreamsim} and apply K-means clustering to remove highly similar samples. For the remaining images, we extract the main object mask using BiRefNet \cite{zheng2024bilateral}.

Next, we generate a base prompt from the real image using Qwen2.5-VL \cite{bai2025qwen2}. This prompt is then intentionally perturbed using ChatGPT to induce common low-quality cases relevant to residential interior AIGC. Rather than aiming to exhaust all possible real-world distortions, this process is designed to cover a practically important subset of frequent failure modes. This process yields nine prompts of varying quality for each object.

Finally, we use the nine prompts and corresponding object masks to generate images with four inpainting/editing models: SD1.5-Inpaint, SDXL-BrushNet \cite{brushnet}, FLUX-Inpaint, and FLUX EasyControl \cite{zhang2025easycontrol}. These models were selected because they support stable large-scale interior completion/editing and cover two major generator families, SD and FLUX. In addition to targeted prompt perturbations, the generated results also contain naturally occurring artifacts and inconsistencies from these pipelines. For each real image, this process yields $9 \times 4 = 36$ AI-generated images. SA-BENCH is thus constructed in an inpainting/mask-based setting, which is practically important but narrower than fully unconstrained text-to-image synthesis.

\subsubsection{Annotation protocol}
\begin{figure}[t]
    \centering
    \includegraphics[width=0.9\columnwidth]{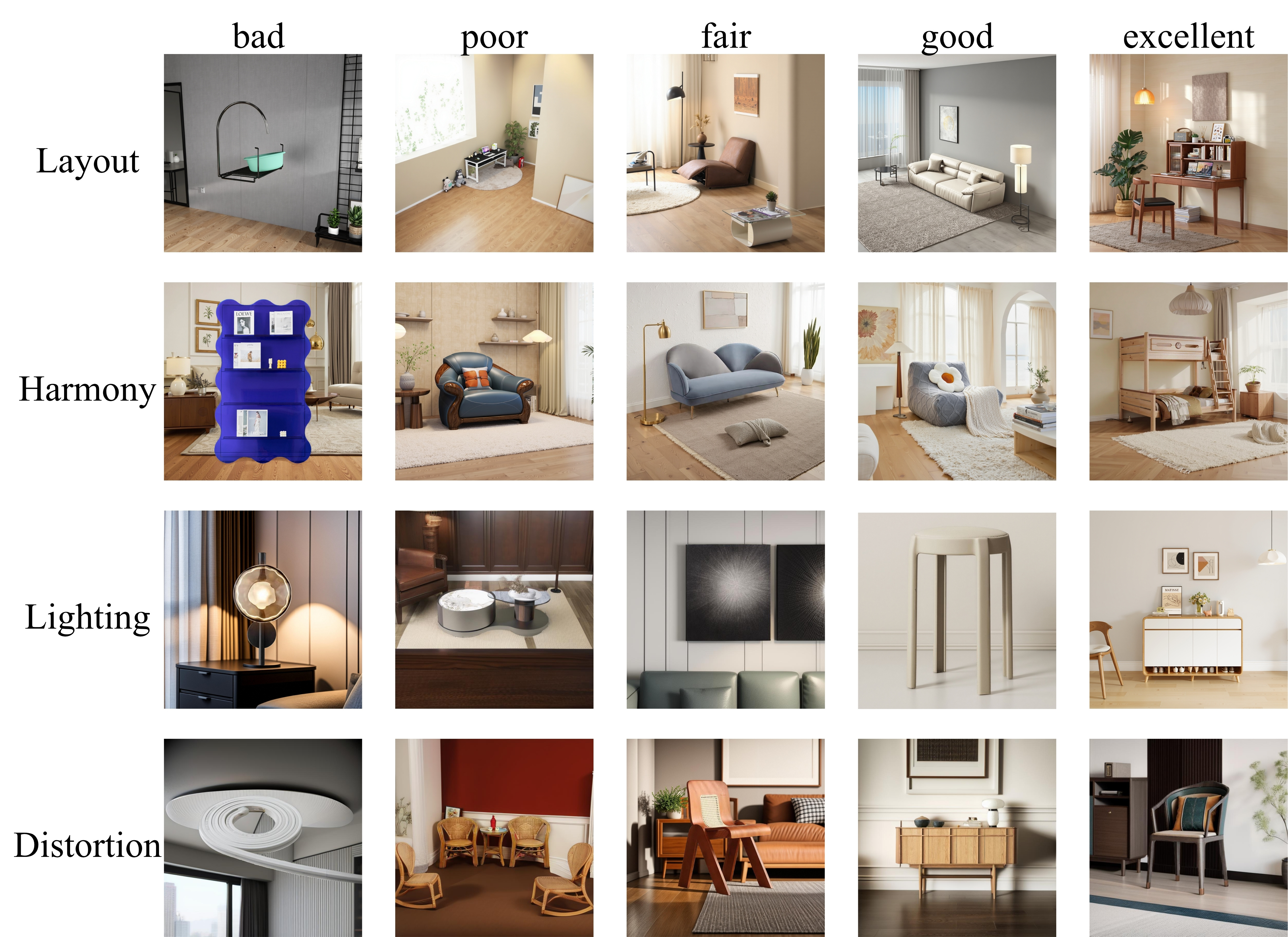}
    \caption{
        \textbf{Sample annotation examples from the SA-Bench.} For each quality dimension (Layout, Harmony, Lighting, and Distortion), five representative examples are presented, spanning quality levels from \emph{bad} (1) to \emph{excellent} (5). These examples serve as crucial visual guidelines for human annotators, ensuring consistent and high-quality scoring throughout our benchmark.
    }
    \label{fig:sa_benchmark_examples}
\end{figure}

\begin{figure}[t]
    \centering
    \includegraphics[width=0.9\columnwidth]{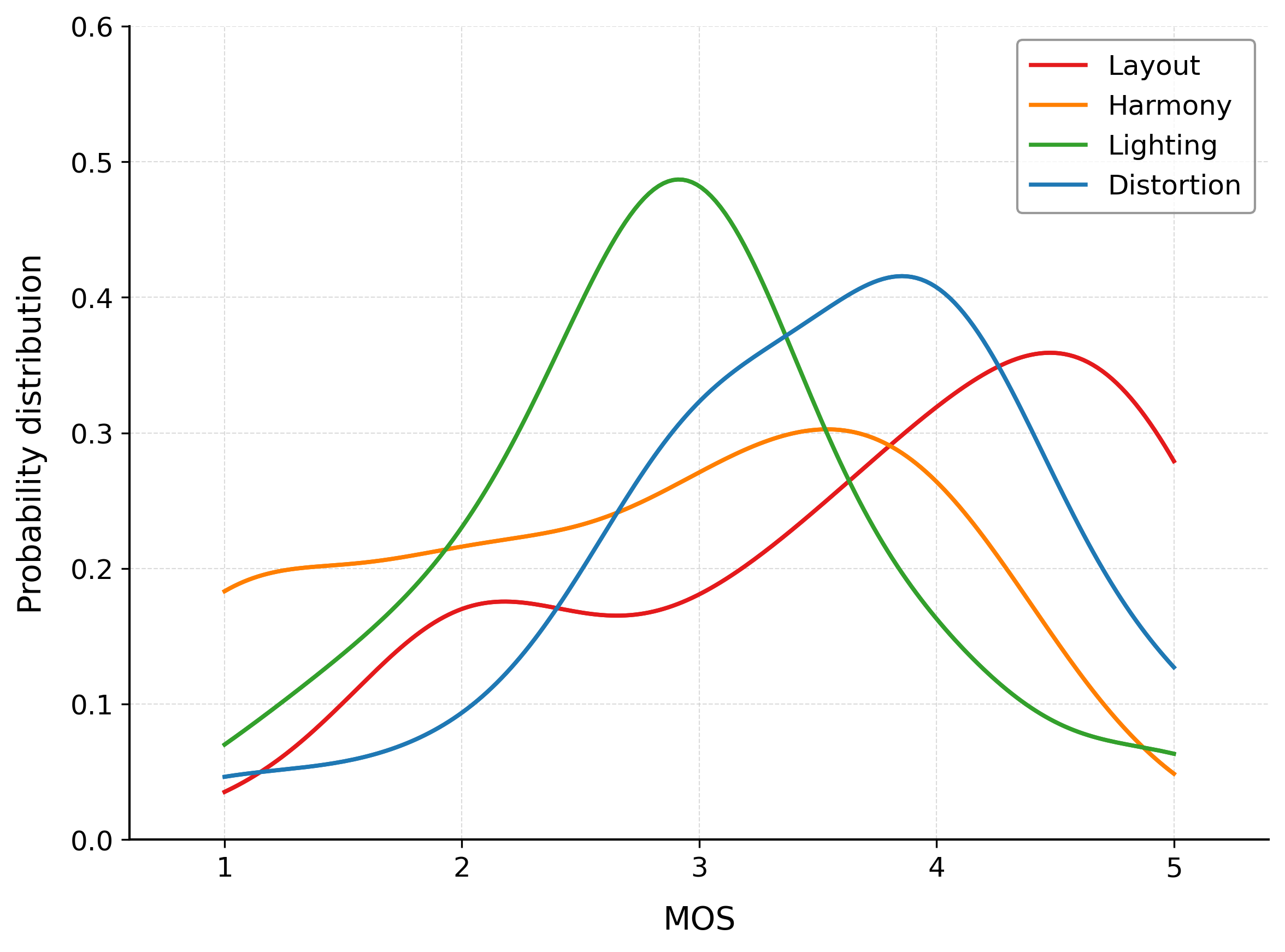}
    
    \caption{
        \textbf{MOS Distribution on SA-Bench.} The plot shows the probability distribution of Mean Opinion Scores (MOS) for each of the four dimensions (Layout, Harmony, Lighting, and Distortion), illustrating the range and concentration of scores from 1 (bad) to 5 (excellent).
    }
    
    \label{fig:mos_distribution}
\end{figure}

Each image is independently rated on a 1--5 scale by 1--5 professionally trained interior designers, with most images receiving five independent ratings. To ensure annotation reliability with minimal reduced subjective bias, SA-BENCH follows a rigorous pipeline comprising small-scale pre-annotation, large-scale expert labelling, and multiple rounds of post-hoc validation/acceptance by dedicated reviewers. Figure~\ref{fig:sa_benchmark_examples} illustrates representative examples for different quality levels across the four dimensions, while Figure~\ref{fig:mos_distribution} shows the resulting score distributions. 
The four dimensions are defined as follows:

\begin{itemize}
    \item Layout: Spatial arrangement of key elements, their relative positions, and object counts. Typical problems include excessive emptiness, clutter, and poor placement.
    \item Harmony: Stylistic coherence, color compatibility, and overall visual consistency. Typical problems include color clashes, mismatched design styles, and generally unattractive appearance.
    \item Lighting: Quality of illumination, shadow behavior, and realism of light sources. Typical problems include unnatural lighting, incorrect or inconsistent shadows, and implausible light sources.
    \item Distortion: Geometric or semantic deformation of furnishings/fixtures and the realism of materials. Typical problems include warped or distorted backgrounds, shape deformation, and unconvincing materials.
\end{itemize}

\subsubsection{Cleaning and reliability}
Acceptance auditing. We audit 10\% of annotations; batches falling below 85\% accuracy are returned for re-annotation.

Score aggregation. We compute the MOS to aggregate $N_i$ ratings ($s_{ij}$) and reduce annotator bias:
\begin{equation}
    \text{MOS}_{i,d} = \frac{1}{N_i} \sum_{j=1}^{N_i} s_{ij}
\end{equation}

Rater reliability. We compute the SRCC between annotator scores and the MOS. Annotators with $\text{SRCC} < 0.6$ are flagged for review.

Outlier mitigation. We standardize ratings (z-score) to identify outliers. Ratings with an absolute z-score $> 2$ are replaced by the mean, and the MOS is recomputed. This stabilizes aggregation against clear annotation noise, although it may also reduce part of the valid subjective variance.

Finally, SA-BENCH contains 17,753 images and 50,476 annotations, providing the first large-scale benchmark for indoor spatial aesthetic IQA (Table~\ref{tab:dataset_stats}).

\begin{table}[t]
\centering
\caption{Statistics of the SA Benchmark Dataset.}
\label{tab:dataset_stats}
\begin{tabular}{@{}lrrrr@{}} 
\toprule
\textbf{Dimension }  & \textbf{Annotations} & \textbf{Train} & \textbf{Test} & \textbf{Total} \\
\midrule
Layout & 23,615 & 4,444 & 1,098 & 5,542 \\
Harmony & 19,615 & 5,392 & 1,348 & 6,740 \\
Lighting & 4,546 & 2,596 & 649 & 3,245 \\
Distortion & 2,700 & 1,780 & 446 & 2,226 \\
\midrule
\textbf{Total} & \textbf{50,476} & 14,212 & 3,541 & \textbf{17,753} \\
\bottomrule
\end{tabular}
\end{table}

\subsection{SA-IQA}
\subsubsection{IQA model}
\paragraph{Model Architecture and Fine-Tuning}
Our SA-IQA model is built upon the Ovis2.5 \cite{Lu2025Ovis2.5} MLLM backbone, chosen for its strong high-resolution and spatial sensitivity. We perform supervised fine-tuning (SFT) on our SA-BENCH train dataset to assess four aesthetic dimensions: layout, harmony, lighting, and distortion. As an image-level quality and reward model, SA-IQA primarily relies on visual evidence, while dimension-conditioned prompts provide lightweight guidance for structured judgments. As shown in Figure~\ref{fig:method_overview}, the model processes an image and a dimension-specific text query (Type 4, Table~\ref{tab:prompts_data}) that explicitly defines the target dimension and its evaluation criteria. We attach a dimension tag (e.g., \texttt{<Layout>}) within the prompt. The model outputs a structured response with a specific rating word, such as \texttt{excellent}, \texttt{good}, \texttt{fair}, \texttt{poor}, or \texttt{bad}.

\paragraph{Score Computation}
To convert this categorical text rating into a continuous MOS score, we employ a probabilistic method. This process involves extracting the top-5 probabilities for the five predefined rating words (e.g., \texttt{excellent} to \texttt{bad}) from the MLLM's final quality word token. After softmax normalization, we compute the expected value over a 1-5 scale (e.g., \texttt{excellent}$\rightarrow$5, \texttt{bad}$\rightarrow$1) to derive the final score within the range $[1,5]$.

\subsubsection{Multidimensional Score Fusion}
Downstream applications, such as reinforcement learning, often require a single composite quality score rather than multiple dimensional assessments. Therefore, we fuse the four dimension-specific scores into a single composite quality score by learning optimal weights using a Bradley–Terry (BT) \cite{bradley1952rank} rank-likelihood objective. We construct a rank dataset of 750 image pairs, where annotators provide pairwise preferences (``Image A better,'' ``Image B better,'' or ``Tie'') for overall spatial aesthetics. For each pair of images ($I_A, I_B$), we first compute their respective four-dimensional score vectors, $\mathbf{x}_A$ and $\mathbf{x}_B$, using our SA-IQA model.

We then learn a set of optimal fusion weights $\mathbf{w}$ to compute a final scalar score $S = \mathbf{x}^{\top}\mathbf{w}$. To find these weights, we fit a Bradley–Terry (BT) model. This model defines the probability of $I_A$ being preferred over $I_B$ as a function of their weighted score difference: $P(A > B) = \sigma(S_A - S_B) = \sigma((\mathbf{x}_A - \mathbf{x}_B)^{\top}\mathbf{w})$.

We optimize $\mathbf{w}$ by minimizing the negative log-likelihood (BT-loss) between the model's predicted probabilities and the human-annotated preference labels:
\begin{equation}
\mathcal{L}(\mathbf{w}) = -\,\mathbb{E} \Big[
    y \log \sigma(\Delta\mathbf{x}^{\top}\mathbf{w})
    + (1-y) \log \big(1-\sigma(\Delta\mathbf{x}^{\top}\mathbf{w})\big)
\Big]
\label{eq:bt_loss}
\end{equation}
where $\Delta\mathbf{x} = \mathbf{x}_A - \mathbf{x}_B$ represents the difference between the multidimensional score vectors of the two images in a pair. $y \in \{0,1\}$ is the ground-truth preference label (e.g., $y=1$ if $I_A$ is preferred over $I_B$, $y=0$ otherwise), derived from the human annotations. $\sigma(\cdot)$ is the sigmoid function. Minimizing this loss yields the optimal fusion weights $\mathbf{w}^*$.

The final fused score is then computed as a weighted sum of the dimension scores:
\begin{equation}
\text{Score} = \mathbf{x}^{\top}\mathbf{w}^* = \sum_{i=1}^{n} x_i w_i^*
\label{eq:fusion_score}
\end{equation}
where $x_i$ are the individual dimension scores and $w_i^*$ are the learned optimal weights.
\section{Experiments}
\label{sec:experiments}

\begin{table*}[t]
\centering
\caption{PLCC/SRCC performance comparison on SA-BENCH. Best results are \textbf{bolded}, and second-best results are \underline{underlined}.}
\label{tab:iqa_comparison}
\resizebox{\textwidth}{!}{
% \scriptsize % 缩小字体（\scriptsize < \footnotesize < \small）
\tiny % 将字体设置为 \tiny，比 \scriptsize 更小
\begin{tabular}{lccccc}
\toprule
\textbf{Method Category} & \textbf{Layout} & \textbf{Harmony} & \textbf{Lighting} & \textbf{Distortion} & \textbf{Overall} \\ 
\midrule
\multicolumn{6}{l}{\textbf{Traditional NR-IQA}} \\ 
\midrule
MUSIQ\cite{Ke2021MUSIQ} & 0.121 / 0.138 & 0.007 / -0.017 & 0.133 / 0.228 & 0.019 / -0.001 & 0.057 / 0.084 \\
DBCNN\cite{Zhang2018DBCNN} & 0.162 / 0.173 & -0.006 / -0.007 & \textbf{0.198 / 0.197} & 0.105 / 0.096 & 0.082 / 0.1 \\
CLIPIQA\cite{wang2023clipiqa} & \textbf{0.197 / 0.230} & \underline{0.097 / 0.101} & 0.062 / 0.084 & \underline{0.113 / 0.128} & 0.118 / 0.135 \\
MANIQA\cite{Yang2022MANIQA} & \underline{0.194 / 0.210} & 0.096 / 0.105 & 0.125 / 0.147 & 0.071 / 0.073 & \underline{0.118 / 0.139} \\
HyperIQA\cite{su2020blindly} & 0.179 / 0.217 & \textbf{0.135 / 0.136} & \underline{0.161 / 0.181} & \textbf{0.163 / 0.131} & \textbf{0.144 / 0.165} \\
\midrule
\multicolumn{6}{l}{\textbf{Deep Learning-based NR-IQA}} \\ 
\midrule
Q-Insight\cite{qinsight} & 0.053 / 0.060 & 0.080 / 0.093 & 0.286 / 0.286 & 0.067 / 0.032 & 0.083 / 0.089 \\
Compare2Score\cite{zhu24adaptive} & 0.160 / 0.162 & 0.074 / 0.075 & 0.274 / 0.288 & 0.028 / 0.011 & 0.088 / 0.115 \\
DeQA-Score\cite{you25deqa} & \underline{0.201 / 0.174} & 0.139 / 0.105 & \textbf{0.305 / 0.351} & 0.025 / 0.006 & 0.111 / 0.141 \\
Q-SiT\cite{QSIT} & 0.141 / 0.190 & 0.098 / 0.077 & 0.207 / 0.219 & \textbf{0.184 / 0.131} & 0.115 / 0.140 \\
Q-Align-quality\cite{qalign} & 0.057 / 0.101 & \underline{0.183 / 0.187} & \underline{0.299 / 0.294} & 0.088 / 0.074 & 0.110 / 0.128 \\
Q-Align-aesthetics\cite{qalign} & 0.075 / 0.082 & \textbf{0.318 / 0.320} & 0.116 / 0.120 & 0.001 / 0.026 & \underline{0.133 / 0.142} \\
Q-Eval-Score\cite{zhang2025q} & \textbf{0.265 / 0.266} & 0.149 / 0.142 & 0.036 / 0.042 & \underline{0.144 / 0.153} & \textbf{0.162 / 0.168} \\
\midrule
\multicolumn{6}{l}{\textbf{Commercial MLLMs}} \\ 
\midrule
gpt-4o & 0.017 / -0.009 & 0.537 / 0.542 & \textbf{0.064 / 0.038} & \underline{0.233 / 0.149} & 0.312 / 0.298 \\
gpt-5 & \underline{0.185 / 0.147} & 0.483 / 0.490 & -0.031 / -0.054 & 0.204 / 0.177 & 0.318 / 0.308 \\
claude35\_sonnet & -0.001 / -0.032 & \underline{0.542 / 0.552} & -0.179 / -0.197 & 0.058 / -0.006 & 0.326 / 0.304 \\
qwen-vl-max & 0.088 / 0.067 & 0.526 / 0.528 & -0.037 / -0.058 & 0.091 / 0.078 & \underline{0.362 / 0.359} \\
gemini-2.5-pro & \textbf{0.246 / 0.181} & \textbf{0.594 / 0.603} & \underline{-0.023 / -0.043} & \textbf{0.252 / 0.225} & \textbf{0.414 / 0.393} \\
\midrule
\multicolumn{6}{l}{\textbf{MLLMs (SFT-based)}} \\ 
\midrule
Ovis2-5-2B\cite{Lu2025Ovis2.5} & 0.812 / 0.807 & 0.890 / 0.888 & 0.692 / 0.681 & 0.556 / 0.520 & 0.848 / 0.844 \\
mPLUG-Owl3-7B\cite{Ye2024mPLUGOwl3} & 0.784 / 0.773 & 0.831 / 0.822 & 0.662 / 0.675 & 0.456 / 0.421 & 0.810 / 0.810 \\
GLM-4-1V-9B\cite{zeng2024chatglm} & \underline{0.831 / 0.821} & 0.848 / 0.834 & 0.690 / 0.690 & 0.532 / 0.503 & 0.836 / 0.842 \\
InternVL3-5-8B\cite{wang2025internvl3} & 0.774 / 0.769 & 0.885 / 0.882 & 0.702 / 0.693 & 0.539 / 0.462 & 0.837 / 0.829 \\
Qwen3-VL-2B\cite{qwen3vl2025} & 0.808 / 0.802 & 0.886 / 0.884 & 0.684 / 0.687 & 0.481 / 0.439 & 0.841 / 0.836 \\
Qwen3-VL-4B\cite{qwen3vl2025} & 0.821 / 0.807 & \underline{0.892 / 0.891} & \underline{0.712 / 0.701} & \underline{0.558 / 0.488} & \underline{0.853 / 0.849} \\
Qwen3-VL-8B\cite{qwen3vl2025} & 0.814 / 0.804 & 0.892 / 0.889 & 0.696 / 0.705 & 0.529 / 0.506 & 0.849 / 0.846 \\
SA-IQA(Ours) & \textbf{0.831 / 0.822} & \textbf{0.896 / 0.895} & \textbf{0.724 / 0.694} & \textbf{0.657 / 0.596} & \textbf{0.864 / 0.860} \\
\bottomrule
\end{tabular}
}
\end{table*}

\subsection{Experimental Setup}
\label{sec:setup}
Training details: We fine-tune all MLLMs via supervised fine-tuning (SFT) using AdamW with a learning rate of $2 \times 10^{-5}$ for 3 epochs. Only the LLM component is updated, while the ViT and Aligner remain frozen. Training is conducted on 4 NVIDIA H20 GPUs with a per-GPU batch size of 2; gradient accumulation is used to reach an effective global batch size of 256.

Datasets and metrics: SA-BENCH is used for both training and evaluation, with a 4:1 train/test split matched in data distribution. The split is performed at the scene level to avoid content leakage, ensuring that all images from the same scene remain in the same partition. We report results on the four dimensions—layout, harmony, lighting, and distortion—as well as an overall score on the pooled test set. Evaluation uses PLCC and SRCC between predicted scores and human MOS.

\begin{table*}[t]
\centering
\caption{PLCC/SRCC performance on SA-BENCH for different prompt types. Best results are \textbf{bolded}, and second-best are \underline{underlined}.}
\label{tab:prompts_data}
\resizebox{\textwidth}{!}{
\footnotesize % 缩小字体（\scriptsize < \footnotesize < \small）
\begin{tabular}{lcccccc}
\toprule
\textbf{Prompt Type} & \textbf{Layout} & \textbf{Harmony} & \textbf{Lighting} & \textbf{Distortion} & \textbf{Overall} & \textbf{Description Stype} \\
\midrule
Type 1 & 0.829/0.820 & 0.894/0.891 & 0.717/0.697 & \underline{0.575/0.552} & 0.857/0.855 & Concise \\
Type 2 & 0.825/0.816 & 0.895/0.893 & \textbf{0.729/0.711} & 0.572/0.564 & 0.858/0.854 & Concise+ \\
Type 3 & \underline{0.830/0.823} & \textbf{0.898/0.896} & 0.717/0.702 & 0.567/0.555 & \underline{0.859/0.856} & Detailed \\
Type 4 & \textbf{0.831/0.822} & \underline{0.896/0.895} & \underline{0.724/0.694} & \textbf{0.657/0.596} & \textbf{0.864 / 0.860} & Expert-Aware \\
\bottomrule
\end{tabular}
}
\end{table*}

\begin{table*}[t]
\centering
\caption{PLCC/SRCC performance on SA-BENCH for different MLLM sizes. Best results are \textbf{bolded}, and second-best are \underline{underlined}.}
\label{tab:model_scaling}
\resizebox{\textwidth}{!}{ % 使用 \resizebox 确保表格占据双栏宽度
%\scriptsize % 缩小字体（\scriptsize < \footnotesize < \small）
\tiny % 将字体设置为 \tiny，比 \scriptsize 更小
\begin{tabular}{lcccccc}
\toprule
\textbf{Method Category} & \textbf{Layout} & \textbf{Harmony} & \textbf{Lighting} & \textbf{Distortion} & \textbf{Overall} \\ 
\midrule
Qwen3-VL-2B & 0.808 / 0.802 & \underline{0.886 / 0.884} & 0.684 / 0.687 & 0.481 / 0.439 & 0.841 / 0.836 \\
Qwen3-VL-4B & 0.821 / 0.807 & 0.892 / 0.891 & \underline{0.712 / 0.701} & \underline{0.558 / 0.488} & \underline{0.853 / 0.849} \\
Qwen3-VL-8B & 0.814 / 0.804 & 0.892 / 0.889 & 0.696 / 0.705 & 0.529 / 0.506 & 0.849 / 0.846 \\
Ovis2-5-2B & 0.812 / 0.807 & 0.890 / 0.888 & 0.692 / 0.681 & 0.556 / 0.520 & 0.848 / 0.844 \\
Ovis2-5-9B & \textbf{0.831 / 0.822} & \textbf{0.896 / 0.895} & \textbf{0.724 / 0.694} & \textbf{0.657 / 0.596} & \textbf{0.864 / 0.860} \\
\bottomrule
\end{tabular}
}
\end{table*}

\subsection{IQA Comparisons}
\label{sec:iqa_compare}
We systematically evaluate various IQA approaches on SA-BENCH, categorized into: (i) Traditional NR-IQA methods, (ii) Deep Learning-based NR-IQA models, (iii) Commercial MLLMs, and (iv) our SA-IQA framework utilizing Supervised Fine-Tuning (SFT-based MLLMs). 
Table~\ref{tab:iqa_comparison} summarizes their correlation performance against human MOS across these method groups.

Traditional NR-IQA: These methods exhibit limited performance across all four dimensions of interior spatial aesthetics, with correlations typically below 0.20. While CLIP-IQA shows the strongest layout correlation (0.197/0.230) and HyperIQA leads in harmony (0.135/0.136), distortion (0.163/0.131), and overall (0.144/0.165), the significant performance gap to other categories indicates poor generalization from generic IQA to AIGI interior scenes.

Deep Learning-based NR-IQA: While offering improvements over traditional baselines, these models still achieve only moderate correlations on SA-BENCH. Q-Eval-Score leads in layout (0.265/0.266) and overall performance within this group (0.162/0.168). Q-Align-aesthetics excels in harmony (0.318/0.320), DeQA-Score in lighting (0.305/0.351), and Q-SiT in distortion (0.184/0.131). Despite these gains, their absolute correlations remain substantially lower than those of SFT-based VLMs, particularly for lighting and distortion. Notably, many of these baselines are pre-trained on large-scale aesthetic/IQA corpora (e.g., AVA) and thus already possess generic quality awareness; for a fair and category-consistent comparison, we report their performance without further fine-tuning on SA-BENCH.

Commercial MLLMs: Among closed-source commercial systems, gemini-2.5-pro demonstrates the strongest competitive performance, ranking first in multiple dimensions and overall score (0.414/0.393). Qwen-VL-Max is the next strongest overall (0.362/0.359). Notably, all commercial models struggle with lighting, often yielding near-zero or negative correlations, suggesting limitations in modeling photometric realism and shadow nuances in interior environments.

MLLMs (SFT-based): SFT substantially boosts spatial-aesthetics IQA. Our SA-IQA model, leveraging supervised fine-tuning, achieves state-of-the-art results across all dimensions. It delivers the best performance in layout (0.831/0.822), harmony (0.896/0.895), lighting (0.724/0.694), distortion (0.657/0.596), and overall (0.864/0.860). Compared to the strongest non-our baseline (e.g., Qwen3-VL-8B, overall 0.849/0.846), SA-IQA consistently demonstrates significant improvements, particularly in PLCC for layout and lighting, while maintaining competitiveness in distortion.

These results reveal three key insights: (1) traditional and generic IQA models are ineffective for the multi-dimensional complexities of interior AIGI; (2) general-purpose commercial MLLMs capture stylistic harmony but fall short in robustly assessing lighting and distortion; and (3) aligning MLLMs through supervised fine-tuning on SA-BENCH, combined with multi-dimensional reward evaluation, leads to substantial and consistent gains, establishing a new state of the art on SA-BENCH.

\subsection{Ablation Studies}
\label{sec:ablation}
\paragraph{Prompt Style Analysis.}
Table~\ref{tab:prompts_data} compares four prompt styles: Type 1 (concise), Type 2 (concise with special characters), Type 3 (detailed natural language), and Type 4 (expert-aware and dimension-specific). Type 4 achieves the best overall correlation, indicating that detailed and professionally aligned instructions are beneficial for stable spatial-aesthetics assessment. Full prompt templates are provided in the supplementary material.

\paragraph{Effect of model size.}
Table~\ref{tab:model_scaling} reports the performance of different model scales, including Qwen3-VL (2B, 4B, 8B) and Ovis2.5 (2B, 9B). Moderate scaling improves performance, but gains saturate for larger generic backbones. Ovis2.5-9B with SA-IQA SFT achieves the best overall result (0.864/0.860), highlighting the importance of combining model capacity with task-specific alignment.

\paragraph{Effect of data scale.}
We further examine data scaling on the under-represented \emph{distortion} dimension by adding 1,500 annotated samples. This targeted augmentation improves distortion PLCC from 0.657 to 0.878 and also slightly improves the overall PLCC from 0.864 to 0.876, showing the value of increasing high-quality supervision in scarce dimensions.

\subsection{Reinforcement Learning with GRPO}
\label{sec:rl_grpo}
To demonstrate the efficacy of SA-IQA as a reward signal, we integrate it into the reinforcement learning framework to optimize a prompt expansion module for generative background-completion model. The optimization process involves two stages: (1) LoRA-based SFT on \texttt{Qwen2.5-VL-7B}, followed by (2) GRPO training initialized from the SFT checkpoint, utilizing SA-IQA as the core reward. Specifically, the reward is the overall SA-IQA score obtained by fusing the four dimension-wise scores (layout, harmony, lighting, and distortion) with the learned BT-based weights. In this GRPO pipeline, we use FLUX-Inpaint (from the FLUX family) as the underlying generation/background-completion model. We configured the GRPO with a group size of 8 and a batch size of 4.

Through this optimization, the average SA-IQA reward significantly increase from 0.70 to 0.86, accompanied by a reduction in standard deviation from 0.12 to 0.06. This quantitative improvement indicates both higher generated quality and enhanced stability. Moreover, a human blind evaluation further confirms the effectiveness of SA-IQA as a reward model, showing a clear improvement of +11 percentage points in win rate compared to the pre-optimisation baseline. The human evaluation is conducted on $N=200$ test cases by three professional interior designers via pairwise blind comparison based on overall spatial preference.

\begin{figure}[t]
    \centering
    \includegraphics[width=\columnwidth]{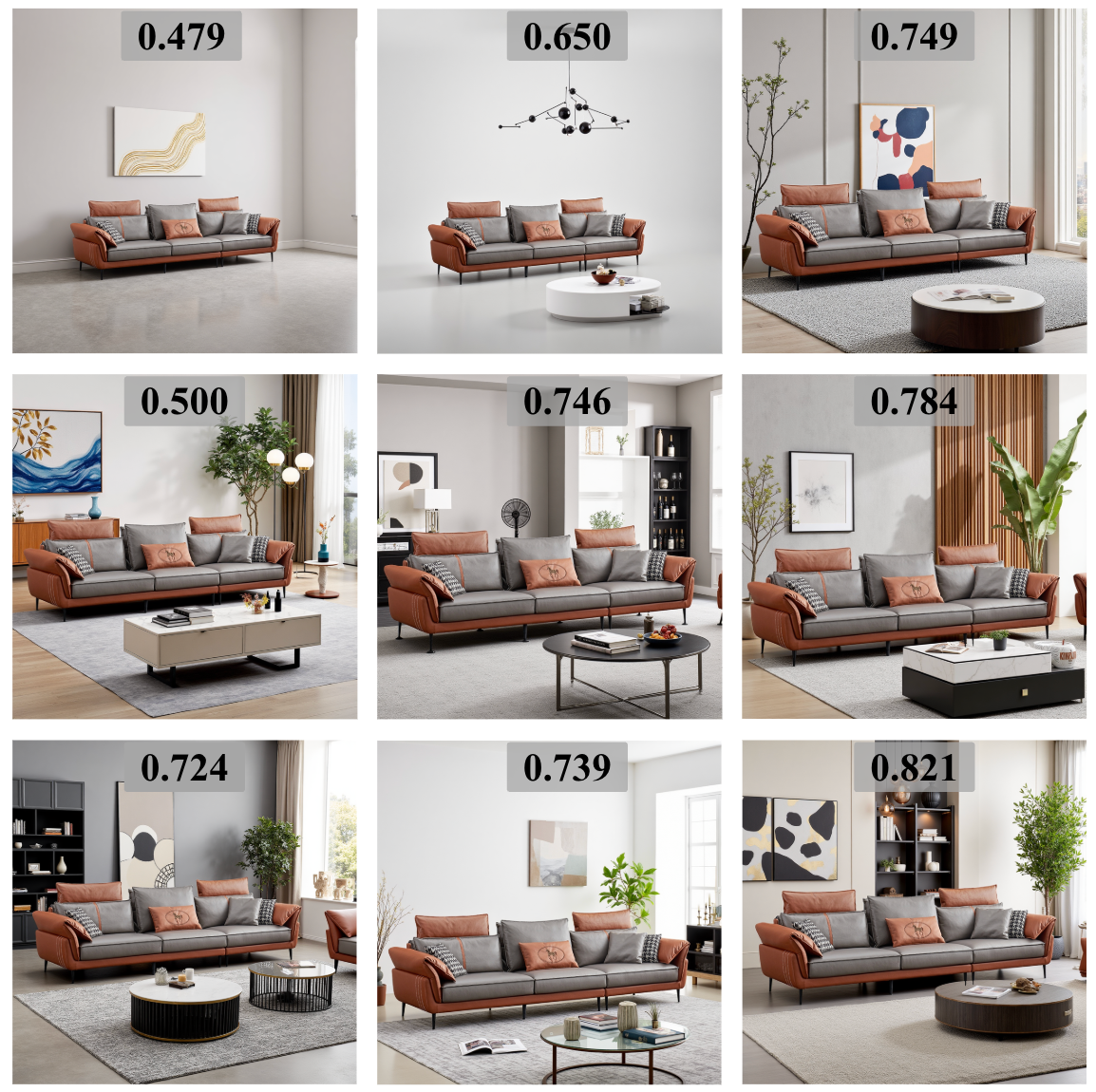}
    \caption{\textbf{Qualitative Visualization of RL Improvement.}
    This figure presents generated background examples from the intermediate training results of our RL process}
    \label{fig:rl_qualitative}
\end{figure}

Figure~\ref{fig:rl_qualitative} visually substantiates these gains. Each row of the figure corresponds to a different training epoch, showcasing the progression of optimization. Within each row, three images are selected from a GRPO group of eight, specifically the 1st, 3rd, and 5th images when sorted by their SA-IQA reward in ascending order. The qualitative comparison clearly illustrates that RL training, guided by SA-IQA, consistently leads to more coherent structural layouts, realistic lighting, and reduced distortions in the generated backgrounds across different epochs.

\subsection{Best-of-N Evaluation}
\label{sec:bon}
To further validate SA-IQA's utility as a generation quality controller, we apply a Best-of-$N$ (BoN) filtering strategy. For each generation prompt, the model samples $N=4$ candidate images. SA-IQA then scores and re-ranks these candidates to select the optimal output. In this Best-of-N pipeline, we use FLUX-Inpaint (FLUX family) as the generation / background inpainting model.

We further compare SA-IQA-based reranking with random selection in the same Best-of-$N$ setting. For each prompt, we either randomly select one candidate or select the highest-scoring candidate according to SA-IQA. The human evaluation is based on pairwise comparison over $N=160$ prompts, where annotators compare the SA-IQA-selected result against the baseline selection. The results show that the SA-IQA-selected image is preferred in 45\% of cases, while random selection is preferred in 31\% of cases, with 24\% judged as comparable. The relatively competitive performance of random selection is mainly due to the limited diversity among candidates generated from the same prompt, which narrows the quality gap in some cases.

SA-IQA effectively sorts candidates based on their spatial aesthetic quality. This re-ranking significantly elevates the final generation quality, demonstrating strong alignment with human preferences. Moreover, a human blind evaluation shows that using SA-IQA as a quality control model yields a substantial improvement, achieving a +14 percentage-point gain in win rate over the baseline without quality-based filtering. These results confirm SA-IQA's reliability in identifying more aesthetically coherent outputs. Additional qualitative results of BoN re-ranking are provided in the supplementary material.
\section{Conclusion}
\label{sec:conclusion}
In this work, we present the first comprehensive study on spatial-aesthetic Image Quality Assessment (IQA) for interior scenes. To this end, we introduce SA-BENCH, the first large-scale, multi-dimensional benchmark comprising 18k images with 50k annotations across four critical dimensions: layout, harmony, lighting, and distortion.

Furthermore, we propose SA-IQA, a novel, reward-ready evaluation framework that achieves state-of-the-art performance. SA-IQA's predictions demonstrate significantly higher PLCC/SROCC correlations with human Mean Opinion Scores (MOS) than existing methods across all four dimensions and overall. We also validate SA-IQA's practical utility as a reward signal. When integrated with GRPO, SA-IQA consistently improves the spatial-aesthetic quality of generated outputs, while Best-of-N selection using our metric further amplifies the output quality. We anticipate this work will inspire and facilitate future research in spatial aesthetics for AI-based interior design.

{
    \small
    \bibliographystyle{ieeenat_fullname}
    \bibliography{main}
}

\clearpage
\setcounter{page}{1}
\maketitlesupplementary

\renewcommand{\thesection}{\Alph{section}}
\renewcommand{\thesubsection}{\thesection.\arabic{subsection}}
\appendix

\section{Overview}
\label{sec:appendix_overview}
This Supplementary Material provides additional details regarding our work on SA-IQA. It is structured as follows: Section~\ref{sec:appendix_prompt_design} elaborates on prompt design strategies. Section~\ref{sec:appendix_dataset_details} details the SA-BENCH dataset, including annotation results and data distributions. Section~\ref{sec:appendix_methodology_details} covers our algorithmic methodology, including data preprocessing, inference scoring, and reinforcement learning with GRPO. Section~\ref{sec:appendix_experimental_details} presents further experimental details on multi-dimensional fusion, training strategies, and failure analysis. Finally, Sections~\ref{sec:iqa_inference_cases} and \ref{sec:limitations_impact} provide additional SA-IQA inference examples and discuss limitations and societal impact, respectively.

\section{Prompt Design}
\label{sec:appendix_prompt_design}

\subsection{SA-IQA Prompts}
\label{sec:sa_iqa_prompts}
The SA-IQA framework utilizes four distinct prompt types, varying from concise to detailed, which were designed for ablation studies. These types include: (1) single short sentences, (2) concise prompts enhanced with special characters, (3) multi-sentence natural language descriptions, and (4) expert-aware prompts incorporating domain-specific terminology. The specific prompts are listed below:

\textbf{Type 1 (Concise)}
\begin{itemize}
    \item Query: \raggedright\texttt{<image>Please evaluate the spatial aesthetic \{dimension\} quality level of this image.} Response: \raggedright\texttt{The spatial aesthetic \{dimension\} quality level of this image is \{rating\_word\}.}\par
\end{itemize}

\textbf{Type 2 (Concise with Special Characters)}
\begin{itemize}
    \item Query: \raggedright\texttt{<image><\{dimension\}>Please evaluate the spatial aesthetic \{dimension\} quality level of this image.} Response: \raggedright\texttt{The spatial aesthetic \{dimension\} quality level of this image is \{rating\_word\}.}\par
\end{itemize}

\textbf{Type 3 (Detailed, General Description)}
\begin{itemize}
    \item Query: \raggedright\texttt{<image><layout>Please evaluate the spatial aesthetic layout quality level of this image. The layout dimension describes the spatial distribution and positional relationships of major elements within a composition. Assess how the layout contributes to the overall organization, structure, and balance of the image.} Response: \raggedright\texttt{The spatial aesthetic layout quality level of this image is \{rating\_word\}.}\par
    \item Query: \raggedright\texttt{<image><harmony>Please evaluate the spatial aesthetic harmony quality level of this image. The harmony dimension emphasizes stylistic consistency, color matching, and overall visual coordination. Consider how well the elements come together to create a unified and pleasing appearance.} Response: \raggedright\texttt{The spatial aesthetic harmony quality level of this image is \{rating\_word\}.}\par
    \item Query: \raggedright\texttt{<image><lighting>Please evaluate the spatial aesthetic lighting quality level of this image. The lighting dimension focuses on the interaction between light and shadow, including the quality of lighting effects and the sense of three-dimensionality. Assess how lighting enhances or affects the depth and overall atmosphere of an image.} Response: \raggedright\texttt{The spatial aesthetic lighting quality level of this image is \{rating\_word\}.}\par
    \item Query: \raggedright\texttt{<image><distortion>Please evaluate the spatial aesthetic distortion quality level of this image. The distortion dimension describes the degree of distortion in shapes or the fidelity of background details. Assess how the distortion impacts the perceived realism and visual quality of the image.} Response: \raggedright\texttt{The spatial aesthetic distortion quality level of this image is \{rating\_word\}.}\par
\end{itemize}

\textbf{Type 4 (Detailed, Expert-Aware Description)}
\begin{itemize}
    \item Query: \raggedright\texttt{<image><layout>Please evaluate the spatial aesthetic layout quality level of this image. The layout dimension describes the spatial distribution, positional relationships, and quantity of major elements within the space. Consider how the layout supports the overall visual order, maintains balance, and enhances the functional aesthetics of the image.} Response: \raggedright\texttt{The spatial aesthetic layout quality level of this image is \{rating\_word\}.}\par
    \item Query: \raggedright\texttt{<image><harmony>Please evaluate the spatial aesthetic harmony quality level of this image. The harmony dimension focuses on stylistic consistency, color coordination, and overall visual cohesion. Examine how well the combination of elements creates a balanced and visually pleasant composition, avoiding clashes or imbalances in style and color.} Response: \raggedright\texttt{The spatial aesthetic harmony quality level of this image is \{rating\_word\}.}\par
    \item Query: \raggedright\texttt{<image><lighting>Please evaluate the spatial aesthetic lighting quality level of this image. The lighting dimension examines the quality of light effects, shadow interactions, and the realism of light sources. Assess how well lighting contributes to the overall depth, mood, and authenticity of the image, emphasizing both natural and artificial lighting scenarios.} Response: \raggedright\texttt{The spatial aesthetic lighting quality level of this image is \{rating\_word\}.}\par
    \item Query: \raggedright\texttt{<image><distortion>Please evaluate the spatial aesthetic distortion quality level of this image. The distortion dimension assesses whether soft furnishings (e.g., cabinets, carpets) or fixed structures (e.g., floors, walls) appear deformed or misaligned. Additionally, evaluate the realism and material accuracy of textures, and judge whether any distortion negatively impacts the overall aesthetic quality of the image.} Response: \raggedright\texttt{The spatial aesthetic distortion quality level of this image is \{rating\_word\}.}\par
\end{itemize}

\begin{table*}[ht]
    \centering
    \caption{\textbf{Definitions and Specific Criteria for Spatial Aesthetic Dimensions.}}
    \label{tab:dimension_criteria_detailed}
    \begin{tabularx}{\textwidth}{@{}lX@{}}
        \toprule
        \textbf{Dimension} & \textbf{Description and Specific Criteria} \\
        \midrule

        \textbf{Layout} & \textbf{Definition:} Spatial arrangement of key elements, their relative positions, and object counts. Typical problems include excessive emptiness, clutter, and poor placement. \\
        & \textbf{Specific Criteria:} All objects exhibit logical positional relationships, avoiding \textit{Floating Objects}, \textit{Unconventional Placement}. Objects should be placed according to design function and common design practices. The quantity of items should be appropriate, avoiding \textit{Excessive Clutter} or \textit{Excessive Emptiness}. For close-up shots where main items like chairs or cabinets occupy a large proportion, fewer auxiliary items are acceptable but not excessive. \\
        \midrule

        \textbf{Harmony} & \textbf{Definition:} Stylistic coherence, color compatibility, and overall visual consistency. Typical problems include color clashes, mismatched design styles, and generally unattractive appearance. \\
        & \textbf{Specific Criteria:} The overall style is unified, avoiding \textit{Inconsistent Style} or obvious incongruity. Color schemes are coordinated and consistent with the design style, avoiding \textit{Discordant Colors}. The image adheres to popular aesthetic trends, avoiding \textit{Poor Aesthetics}, outdated, or unattractive styles. \\
        \midrule

        \textbf{Lighting} & \textbf{Definition:} Quality of illumination, shadow behavior, and realism of light sources. Typical problems include unnatural lighting, incorrect or inconsistent shadows, and implausible light sources. \\
        & \textbf{Specific Criteria:} Prevents \textit{Over-Bright/Unrealistic Light}, such as overexposure due to overly strong or unrealistic light. Ensures \textit{Unrealistic Shadows} are avoided, preventing uniformly bright or dark scenes, and ensuring shadow directions align logically with light sources. Light sources are logically and realistically placed, thereby avoiding an \textit{Implausible Light Source}. \\
        \midrule

        \textbf{Distortion} & \textbf{Definition:} Geometric or semantic deformation of furnishings/fixtures and the realism of materials. Typical problems include warped or distorted backgrounds, shape deformation, and unconvincing materials. \\
        & \textbf{Specific Criteria:} Soft furnishings (various objects) and hard fixtures (e.g., ceilings, walls, floors) are not subjected to \textit{Soft Furnishing Deformation} or \textit{Hard Fixture Deformation}, meaning they are not deformed, warped, or misaligned. Material effects are realistic, with appropriate textures and proportions, avoiding \textit{Unrealistic Materials}, deformation, stretching, or blurriness. \\
        \bottomrule
    \end{tabularx}
\end{table*}

\subsection{Commercial Model Prompts}
\label{sec:commercial_prompts}
To benchmark commercial closed-source models on SA-BENCH, we adapted the expert-aware prompts (Type 4) from SA-IQA. The refined prompts used for querying these models are as follows:

\begin{itemize}
    \item \textbf{Prompt 1 (Layout):} \raggedright\texttt{You are an interior spatial aesthetics evaluation assistant. The input is an image of an interior space. Please evaluate the spatial aesthetic layout quality level of this image. The layout dimension describes the spatial distribution, positional relationships, and quantity of major elements within the space. Consider how the layout supports the overall visual order, maintains balance, and enhances the functional aesthetics of the image. Output only one score from [1,2,3,4,5], where a higher score indicates higher quality. Return in the format \{"score": score\}.}\par
    \item \textbf{Prompt 2 (Harmony):} \raggedright\texttt{You are an interior spatial aesthetics evaluation assistant. The input is an image of an interior space. Please evaluate the spatial aesthetic harmony quality level of this image. The harmony dimension focuses on stylistic consistency, color coordination, and overall visual cohesion. Examine how well the combination of elements creates a balanced and visually pleasant composition, avoiding clashes or imbalances in style and color. Output only one score from [1,2,3,4,5], where a higher score indicates higher quality. Return in the format \{"score": score\}.}\par
    \item \textbf{Prompt 3 (Lighting):} \raggedright\texttt{You are an interior spatial aesthetics evaluation assistant. The input is an image of an interior space. Please evaluate the spatial aesthetic lighting quality level of this image. The lighting dimension examines the quality of light effects, shadow interactions, and the realism of light sources. Assess how well lighting contributes to the overall depth, mood, and authenticity of the image, emphasizing both natural and artificial lighting scenarios. Output only one score from [1,2,3,4,5], where a higher score indicates higher quality. Return in the format \{"score": score\}.}\par
    \item \textbf{Prompt 4 (Distortion):} \raggedright\texttt{You are an interior spatial aesthetics evaluation assistant. The input is an image of an interior space. Please evaluate the spatial aesthetic distortion quality level of this image. The distortion dimension assesses whether soft furnishings (e.g., cabinets, carpets) or fixed structures (e.g., floors, walls) appear deformed or misaligned. Additionally, evaluate the realism and material accuracy of textures, and judge whether any distortion negatively impacts the overall aesthetic quality of the image. Output only one score from [1,2,3,4,5], where a higher score indicates higher quality. Return in the format \{"score": score\}.}\par
\end{itemize}

\begin{figure*}[ht]
    \centering
    \includegraphics[width=\textwidth]{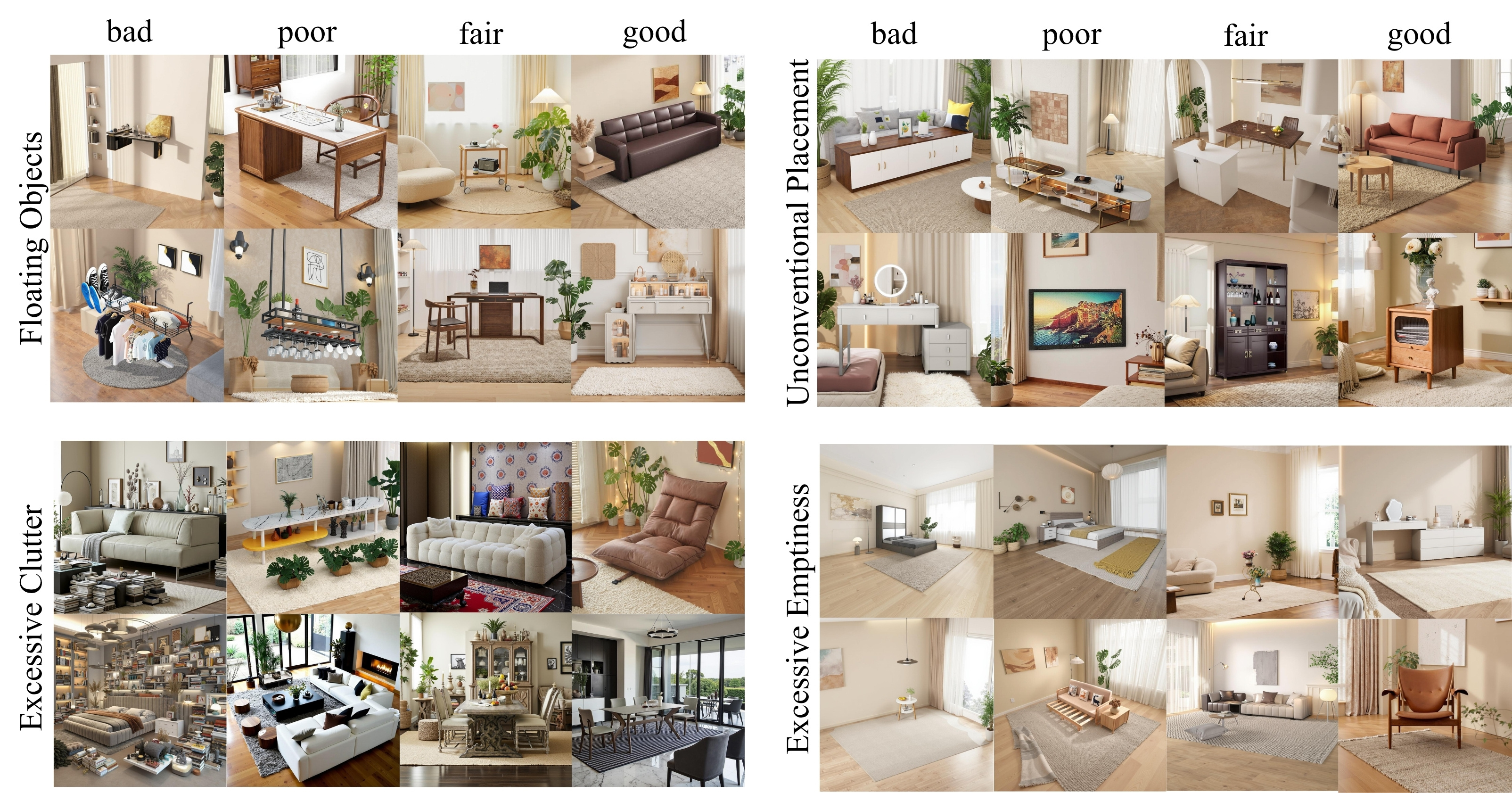}
    \caption{
        \textbf{Layout Dimension Annotation Example.}
        Each row illustrates a specific problem type within the Layout dimension, with image quality incrementally improving from left to right.
    }
    \label{fig:layout_annotation}
\end{figure*}

\begin{figure}[t]
    \centering
    \includegraphics[width=\columnwidth]{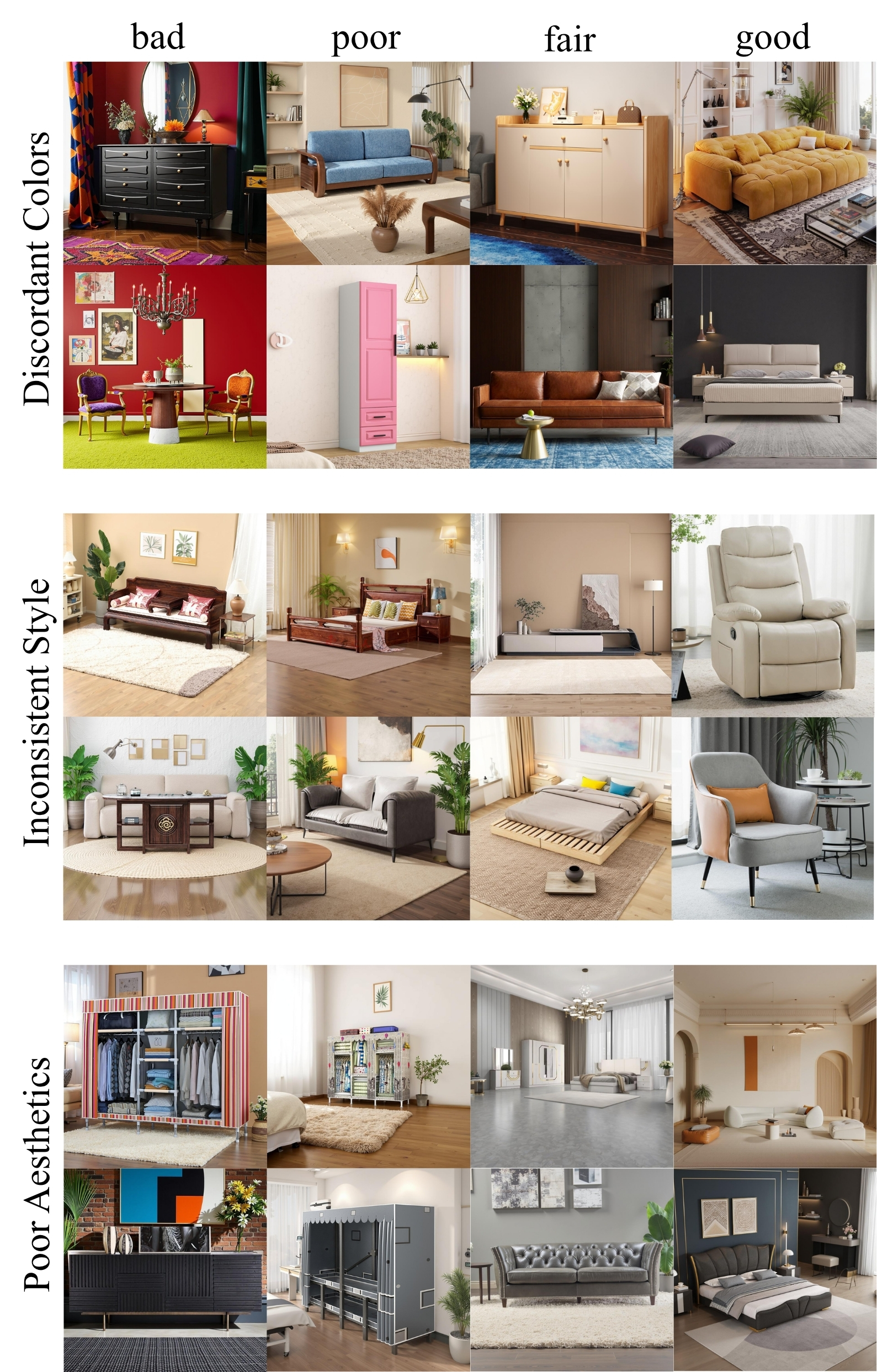}
    \caption{
        \textbf{Harmony Dimension Annotation Example.}
        Similar to Harmony, each row depicts a problem type, showing improving quality from left to right.
    }
    \label{fig:harmony_annotation}
\end{figure}

\begin{figure}[t]
    \centering
    \includegraphics[width=\columnwidth]{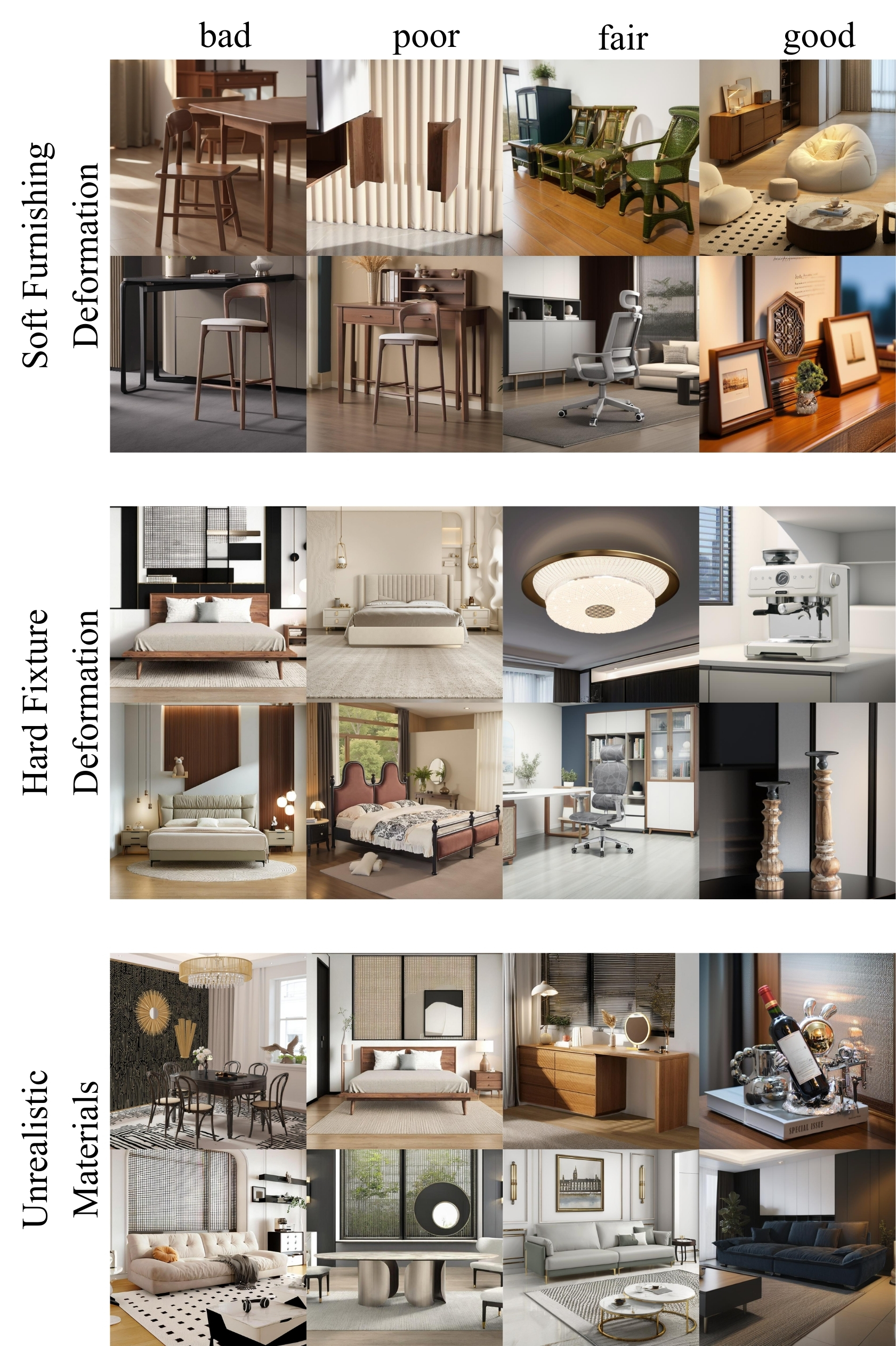}
    \caption{
        \textbf{Distortion Dimension Annotation Example.}
        Each row demonstrates a type of Distortion problem, with image quality increasing from left to right.
    }
    \label{fig:distortion_annotation}
\end{figure}

\section{Dataset Details}
\label{sec:appendix_dataset_details}

\subsection{Annotation Results for Four Dimensions}
\label{sec:annotation_results}

We define four spatial aesthetic dimensions for image quality assessment: Layout, Harmony, Lighting, and Distortion. Their definitions and specific criteria, guiding our annotation process, are summarized in Table \ref{tab:dimension_criteria_detailed}.

\paragraph*{Rating Scale.} Annotations utilize a 5-point Likert scale, where 1 (significant major issue) corresponds to \textit{bad}, 2 (noticeable minor issue) to \textit{poor}, 3 (subtle minor issue) to \textit{fair}, 4 (slight flaw) to \textit{good}, and 5 (no issue) to \textit{excellent}. 
% To enhance reliability and mitigate subjective bias, a Mean Opinion Score (MOS) approach was employed, averaging multiple annotator responses.

\begin{figure}[t]
    \centering
    \includegraphics[width=\columnwidth]{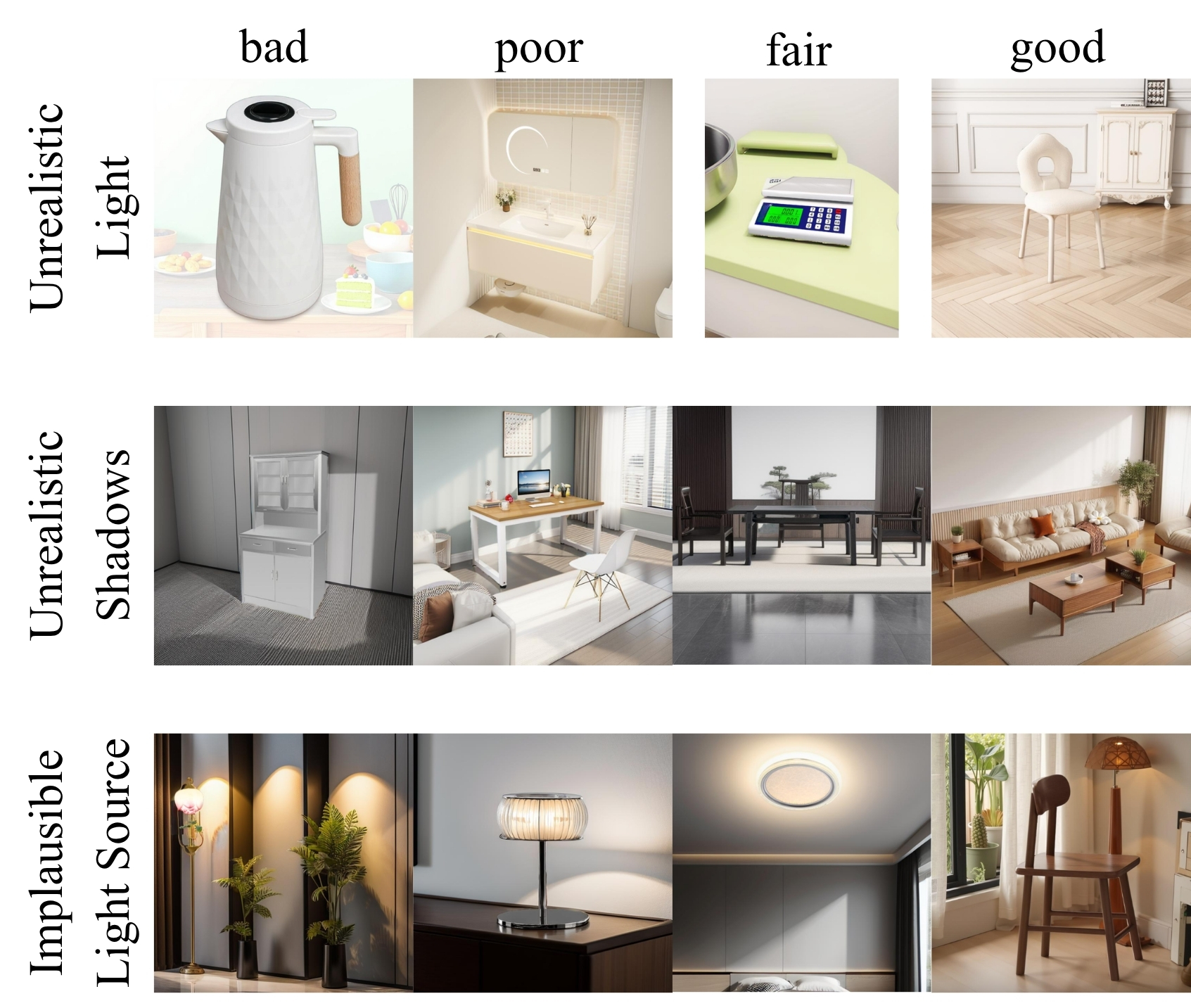}
    \caption{
        \textbf{Lighting Dimension Annotation Example.}
        Image quality for various Lighting issues progressively improves from left to right in each row.
    }
    \label{fig:lighting_annotation}
\end{figure}

\begin{figure}[t]
    \centering
    \includegraphics[width=\columnwidth]{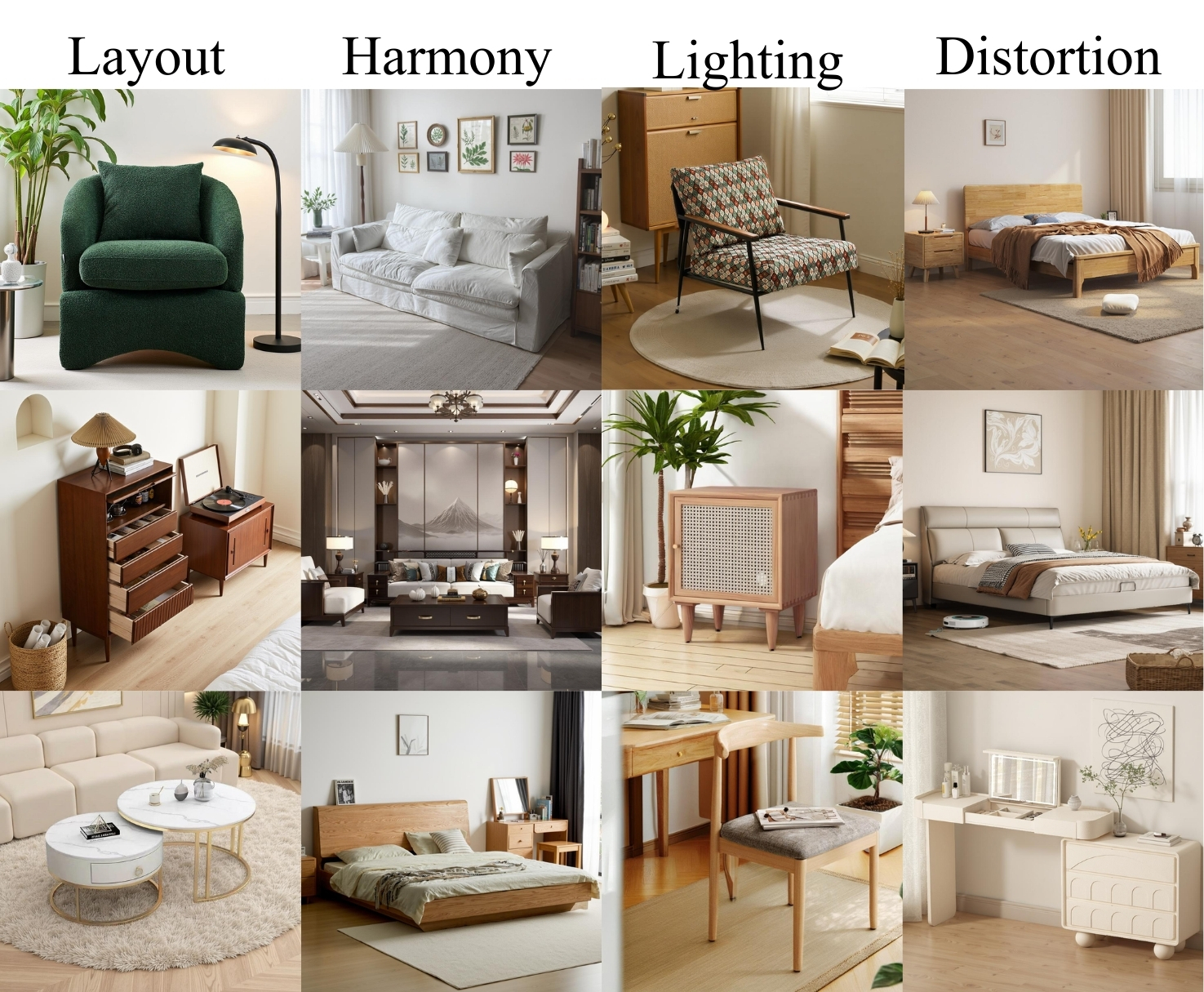}
    \caption{
        \textbf{Excellent Quality Examples Across All Four Dimensions.}
        Illustrative images receiving ``excellent'' quality ratings across all four spatial aesthetic dimensions (Layout, Harmony, Lighting, and Distortion).
    }
    \label{fig:excellent_quality}
\end{figure}

\paragraph*{Visual Annotation Examples.} Visualizations of our annotation results are presented in Figures \ref{fig:layout_annotation} to \ref{fig:excellent_quality}. Specifically, Figures \ref{fig:layout_annotation}, \ref{fig:harmony_annotation}, \ref{fig:lighting_annotation}, and \ref{fig:distortion_annotation} showcase annotation samples for Layout, Harmony, Lighting, and Distortion, respectively, where image quality gradually improves from left to right within each row's problem type. Figure \ref{fig:excellent_quality} provides examples of images that received an ``excellent'' quality rating across all four dimensions.

\begin{figure}[ht]
    \centering
    \includegraphics[width=\columnwidth]{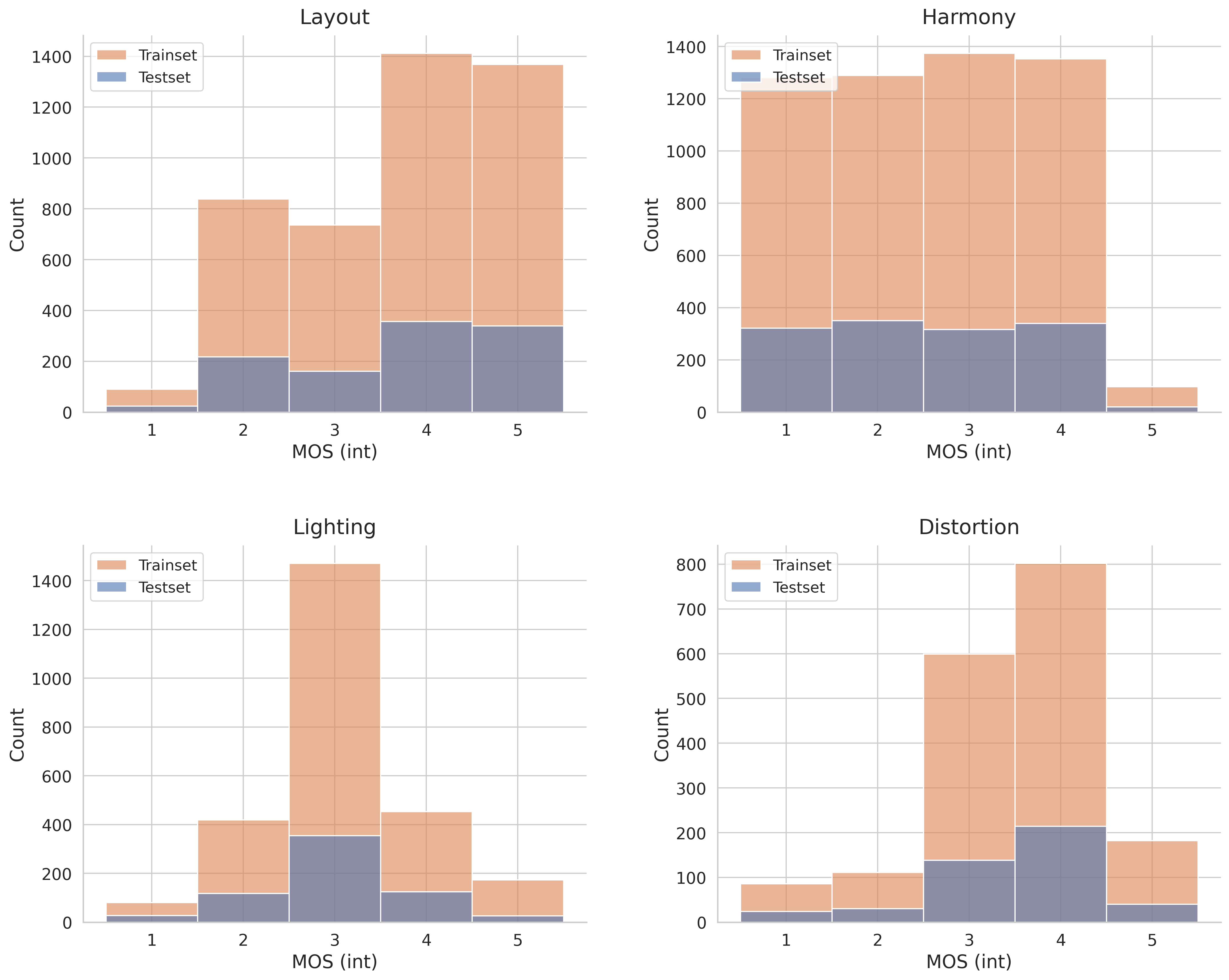}
    \caption{
        \textbf{Distribution of SA-BENCH Across Spatial Aesthetic Dimensions.}
        This figure shows the Mean Opinion Score (MOS) distributions for SA-BENCH's training and testing sets across four spatial aesthetic dimensions: Layout, Harmony, Lighting, and Distortion. Scores are quantized (1-5) for clarity.
    }
    \label{fig:sa_bench_mos_distribution}
\end{figure}

\subsection{Data Distribution}
\label{sec:data_distribution}
The data distribution for the training and testing sets across all four dimensions of SA-BENCH is visualized in Figure~\ref{fig:sa_bench_mos_distribution}. It can be observed that the distributions for both the training and testing sets are consistent for each dimension.

\section{Methodology Details}
\label{sec:appendix_methodology_details}

\subsection{Data Preprocessing}
\label{sec:data_preprocessing}

After data annotation, two key steps were performed for quality control:

\paragraph{Rater Reliability} We evaluate annotator consistency by computing the Spearman's Rank Correlation Coefficient (SRCC) between each individual annotator's scores and the aggregated Mean Opinion Score (MOS). Annotators whose SRCC falls below 0.6 are flagged for further review to ensure data quality. The SRCC is calculated as:
\begin{equation}
    \text{SRCC} = 1 - \frac{6 \sum d_i^2}{n(n^2 - 1)}
\end{equation}
where $d_i$ is the difference between the ranks of corresponding observations, and $n$ is the number of observations.

\paragraph{Outlier Mitigation} We standardize ratings using the z-score to identify and mitigate outliers. Ratings with an absolute z-score $> 2$ are replaced by the mean score for that image, and the MOS is recomputed. The z-score is calculated as:
\begin{equation}
    z = \frac{x - \mu}{\sigma}
\end{equation}
where $x$ is an individual rating, $\mu$ is the mean of all ratings for a given image, and $\sigma$ is the standard deviation of all ratings for that image.

\subsection{Inference Scoring}
\label{sec:inference_scoring}

To bridge the gap between the categorical text ratings generated by the MLLM and a continuous MOS, we employ a probabilistic conversion method. This process involves two main steps:

\paragraph{Softmax Normalization.} We first extract the top-5 probabilities corresponding to the five predefined rating words (e.g., \texttt{excellent}, \texttt{good}, \texttt{fair}, \texttt{poor}, \texttt{bad}) from the MLLM's final quality word token. These raw probabilities are then normalized using the softmax function to ensure they sum to 1. The softmax function is defined as:
\begin{equation}
    P_i = \frac{e^{logits_i}}{\sum_{j=1}^{K} e^{logits_j}}
\end{equation}
where $P_i$ is the normalized probability for rating word $i$, $logits_i$ is the logit for rating word $i$ directly from the MLLM, and $K=5$ is the total number of rating words.

\paragraph{Weighted Sum for Final Score.} After obtaining the normalized probabilities, we compute the expected value by mapping these probabilities to a 1-5 scale (e.g., \texttt{excellent} $\rightarrow$ 5, \texttt{good} $\rightarrow$ 4, \texttt{fair} $\rightarrow$ 3, \texttt{poor} $\rightarrow$ 2, \texttt{bad} $\rightarrow$ 1). This yields the final continuous score within the range $[1, 5]$. The weighted sum formula is:
\begin{equation}
    \text{Score} = \sum_{i=1}^{K} (P_i \times \text{rating\_value}_i)
\end{equation}
where $P_i$ is the normalized probability for rating word $i$, and $\text{rating\_value}_i$ is its corresponding numerical value on the 1-5 scale.

\subsection{Reinforcement Learning with GRPO}
\label{sec:reinforcement_learning_with_GRPO}

We integrate SA-IQA into a reinforcement learning framework to optimize a prompt expansion module for a generative background-completion model. This optimization involves LoRA-based SFT on \texttt{Qwen2.5-VL-7B}, followed by GRPO training (with a group size of 8 and batch size of 4) initialized from the SFT checkpoint, where SA-IQA serves as the core reward signal. 

The calculation formula for GRPO is presented below:
\begin{equation}
\begin{aligned}
    \mathcal{J}_{GRPO}(\theta) = & \mathbb{E}\left[q \sim P(Q), \{o_i\}_{i=1}^G \sim \pi_{\theta_{old}}(O|q)\right] \\
    & \frac{1}{G} \sum_{i=1}^G \frac{1}{|o_i|} \sum_{t=1}^{|o_i|} \Bigg\{ \min \Bigg[ \frac{\pi_\theta(o_{i,t}|q, o_{i,<t})}{\pi_{\theta_{old}}(o_{i,t}|q, o_{i,<t})} \hat{A}_{i,t}, \\
    & \text{clip} \left( \frac{\pi_\theta(o_{i,t}|q, o_{i,<t})}{\pi_{\theta_{old}}(o_{i,t}|q, o_{i,<t})}, 1-\epsilon, 1+\epsilon \right) \hat{A}_{i,t} \Bigg] \\
    & - \beta \mathbb{D}_{KL} \left[\pi_\theta || \pi_{ref}\right] \Bigg\}
\end{aligned}
\label{eq:grpo_objective_correct}
\end{equation}
where $\pi_{\theta}$ and $\pi_{\theta_{old}}$ denote the current policy and the old sampling policy, respectively; $\pi_{ref}$ is the reference policy used for KL divergence regularization with coefficient $\beta$. The expectation is taken over a group of $G$ outputs $\{o_i\}_{i=1}^G$ sampled for a query $q$. $\hat{A}_{i,t}$ represents the advantage value for the $t$-th token of the $i$-th output, which is computed based on the rewards provided by SA-IQA relative to the group average. $\epsilon$ is the clipping parameter used to constrain the policy update.

\begin{table}[t]
    \centering
    \caption{Multi-Dimensional Fusion Comparison on Reward-Benchmark for Ranking Accuracy. \textbf{Bold} indicates the best performance.}
    \label{tab:fusion_comparison}
    \begin{tabular}{lcc}
        \toprule
        \textbf{Eval Dimension} & \textbf{Threshold} & \textbf{Rank Accuracy} \\
        \midrule
        Layout                       & 0.15                     & 0.457                \\
        Harmony                      & 0.27                     & 0.541                \\
        Lighting                     & 0.22                     & 0.449                \\
        Distortion                   & 0.22                     & 0.440                \\
        \midrule
        Equal Weighting (4D)         & 0.37                     & 0.503                \\
        Optimal Weighting (4D)       & 0.32                     & \textbf{0.567}                \\
        \bottomrule
    \end{tabular}
\end{table}

\section{Experimental Details}
\label{sec:appendix_experimental_details}

\begin{table*}[t]
    \centering
    \caption{Impact of Training Strategies (PLCC/SRCC) on SA-IQA Performance.}
    \label{tab:training_impact}
    \resizebox{\textwidth}{!}{
    %\scriptsize % 缩小字体（\scriptsize < \footnotesize < \small）
    \tiny % 将字体设置为 \tiny，比 \scriptsize 更小
    \begin{tabular}{lccccc}
        \toprule
        \textbf{Training Method} & \textbf{Layout} & \textbf{Harmony} & \textbf{Lighting} & \textbf{Distortion} & \textbf{Overall} \\
        \midrule
        LoRA Fine-tuning         & 0.374 / 0.371   & 0.707 / 0.707    & 0.549 / 0.531     & 0.183 / 0.237       & 0.642 / 0.617    \\
        Full Fine-tuning         & 0.831 / 0.822   & 0.896 / 0.895    & 0.724 / 0.694     & 0.657 / 0.596       & 0.864 / 0.860    \\
        \bottomrule
    \end{tabular}
    }
\end{table*}

\subsection{Multi-Dimensional Fusion Comparison}
\label{sec:fusion_comparison}

Our proposed SA-IQA framework, particularly its multi-dimensional fusion strategy, serves as a crucial reward mechanism for optimizing AI-generated content (AIGC) pipelines via reinforcement learning, specifically with GRPO. To validate the effectiveness of our fusion approach, we evaluate its ranking accuracy on a custom reward-benchmark designed to assess overall aesthetic quality across the four dimensions.

We investigate three distinct fusion settings:
\begin{enumerate}
    \item \textbf{Individual Single-Dimension Models:} Each of the four dimensions (Layout, Harmony, Lighting, Distortion) is evaluated independently.
    \item \textbf{Equal-Weighted Fusion (1:1:1:1):} The scores from the four dimensions are combined with equal weighting.
    \item \textbf{Optimal Weighting (via bt-loss):} Fusion is performed using adaptively learned weights, optimized through a \textit{bt-loss} mechanism.
\end{enumerate}

The ``Threshold'' column in Table~\ref{tab:fusion_comparison} represents a dynamically optimized threshold for each ranking method. In our pairwise comparison setup, where two images are evaluated, results can be `win', `lose', or `tie'. This threshold (ranging from 1 to 5) defines that if the absolute difference between the continuous scores of two images falls below it, the comparison is registered as a `tie'.

Table~\ref{tab:fusion_comparison} confirms the efficacy of multi-dimensional fusion. Optimal Weighting (via \textit{bt-loss}) significantly outperforms individual single-dimension models and Equal Weighting in ranking accuracy. Harmony shows stronger individual performance, aligning with its higher weight in optimal fusion. Equal Weighting's underperformance relative to the best individual dimension underscores the critical role of adaptive weighting, as it is unduly influenced by less performant dimensions. The fusion weights are estimated on a validation split by optimizing agreement with human pairwise preferences under the Bradley--Terry formulation. We report the learned relative weights after normalizing harmony to 1.0: layout = 0.353, harmony = 1.000, lighting = 0.380, and distortion = 0.100.

\subsection{Training Strategy Impact}
\label{sec:training_strategy_impact}

We investigate the impact of different training strategies on SA-IQA's performance, specifically comparing LoRA (Low-Rank Adaptation) fine-tuning against full model fine-tuning. Both strategies are applied to the SA-IQA algorithm, fine-tuned on the Ovis2-5-9B model. For LoRA training, the following specific parameters were employed: \texttt{lora\_rank} 32, \texttt{lora\_alpha} 64, \texttt{target\_modules} \texttt{all-linear}, \texttt{freeze\_llm} false, \texttt{freeze\_vit} true, and \texttt{freeze\_aligner} true. All other training parameters were kept consistent with those used for full fine-tuning. The comparative results, presented in Table~\ref{tab:training_impact}, clearly demonstrate that full fine-tuning achieves superior performance compared to LoRA fine-tuning across all evaluated metrics.

\subsection{Failure Analysis of RL Reward Optimization}
\label{sec:failure_analysis}
We observe a reward-hacking failure mode in RL optimization: the reward model may over-score close-up images in which the main product occupies a large portion of the frame and the visible background is limited. Such samples can appear visually clean and less distorted, but often provide insufficient evidence for layout quality and spatial coherence. To reduce this issue, we introduce a detector-based penalty that discourages overly cropped compositions with insufficient background context. This improves robustness in the RL setting, although reward hacking is not fully eliminated.

\section{SA-IQA Inference Case Studies}
\label{sec:iqa_inference_cases}

\begin{figure}[t]
    \centering
    \includegraphics[width=\columnwidth]{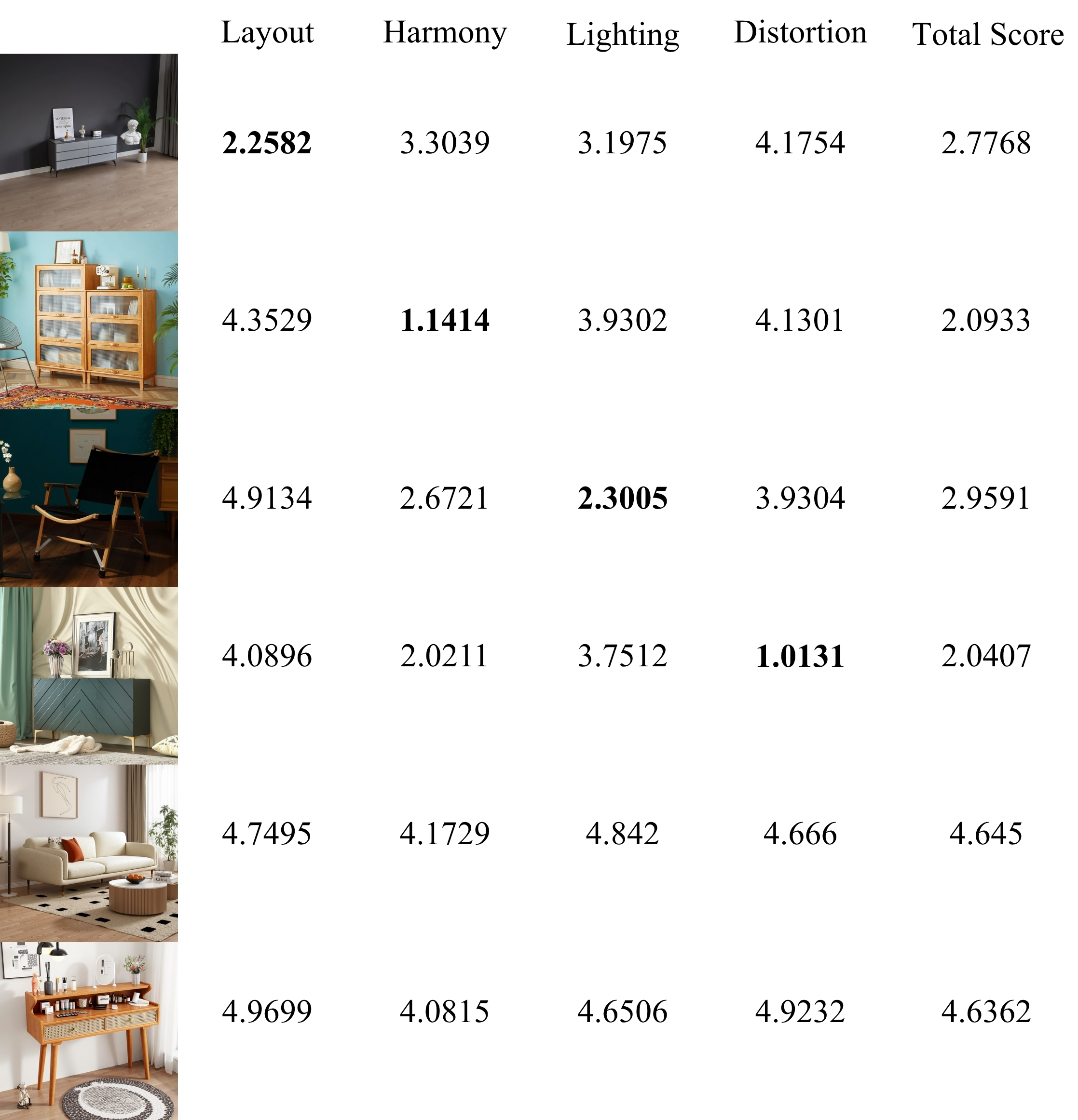}
    \caption{
        \textbf{SA-IQA Inference Examples.}
        This figure demonstrates SA-IQA's multi-dimensional scoring capabilities for interior images. Columns sequentially display scores for Layout, Harmony, Lighting, Distortion, and the overall Total Score. \textbf{Bolded} numerical values within a dimension indicate the presence of identified quality issues.
    }
    \label{fig:sa_iqa_inference_example}
\end{figure}

Figure~\ref{fig:sa_iqa_inference_example} presents illustrative inference results of SA-IQA across diverse interior scene images. For each displayed image, the columns, from left to right, detail the predicted scores for the Layout, Harmony, Lighting, and Distortion dimensions, culminating in an aggregated Total Score. Notably, \textbf{bolded scores} signify dimensions where SA-IQA identifies significant quality deficiencies, providing a clear visual cue for specific aesthetic issues.

\section{Best-of-N Re-ranking Case Studies}
\label{sec:bon_cases}

\begin{figure}[t]
    \centering
    \includegraphics[width=\columnwidth]{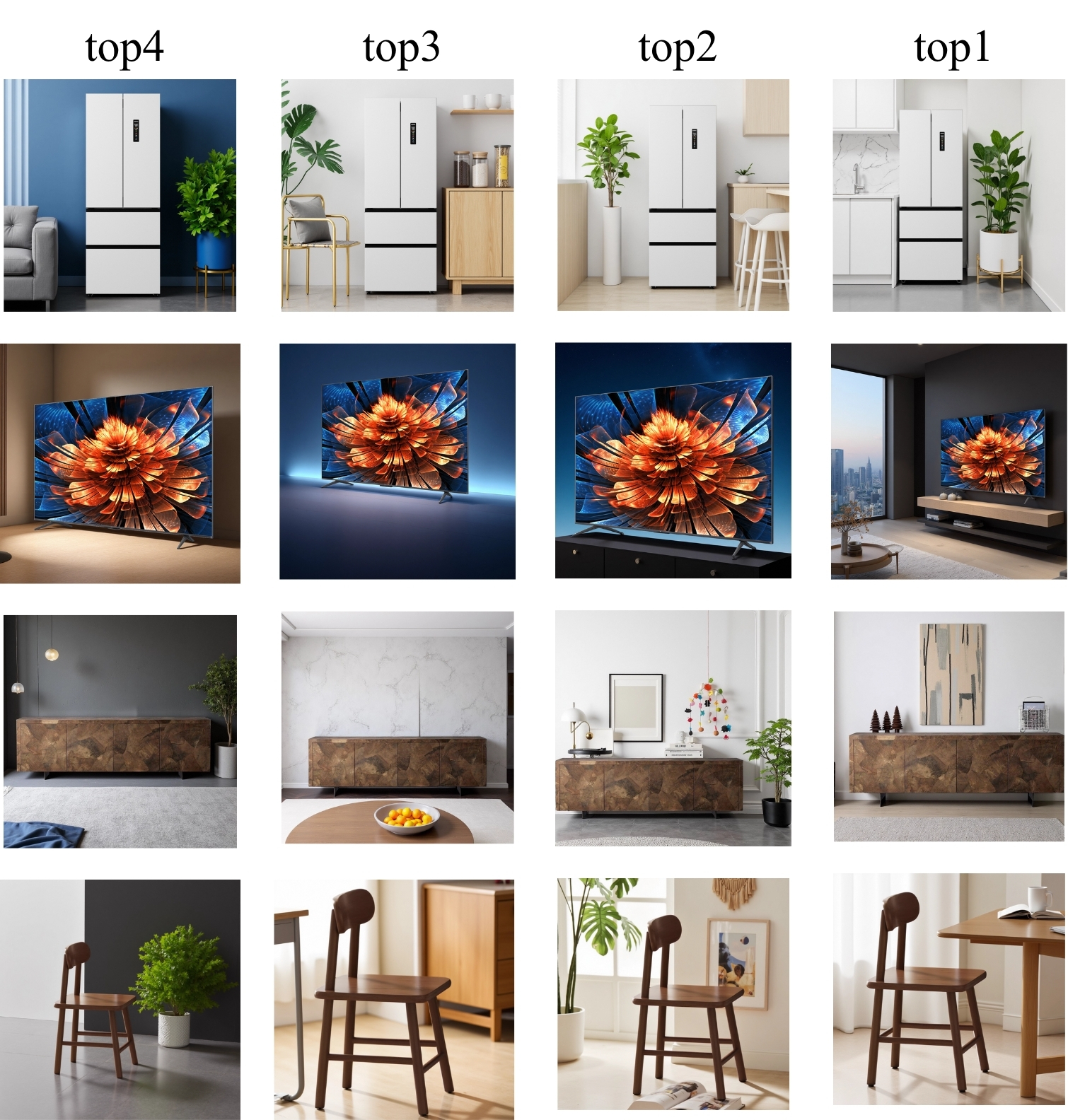}
    \caption{
        \textbf{Best-of-$N$ Re-ranking for Quality Filtering.}
        Each row showcases generated images for a given prompt, sorted from lowest (left) to highest (right) SA-IQA assessed quality. SA-IQA consistently identifies and prioritizes aesthetically superior outputs for interior scenes.
    }
    \label{fig:best_of_n}
\end{figure}

Figure~\ref{fig:best_of_n} provides qualitative case studies of Best-of-$N$ (BoN) selection using SA-IQA as a quality controller. For each prompt, we sample $N=4$ candidate images from the same generation model and then apply SA-IQA to score and re-rank the candidates. In each row, images are ordered from lowest (left) to highest (right) according to the SA-IQA total score, where the rightmost image corresponds to the BoN-selected output. Across diverse indoor scenes, SA-IQA reliably promotes candidates with more coherent layouts, improved style/color harmony, more consistent lighting, and fewer structural distortions or artifacts.

\section{Limitations and Social Impact}
\label{sec:limitations_impact}

\paragraph{Limitations.}
Our SA-IQA, while effective, has several limitations. First, SA-BENCH currently focuses on residential interior furnishing scenes and an inpainting-/mask-based generation setting, which may limit direct generalization to broader spatial domains and fully unconstrained text-to-image synthesis. Second, the current benchmark has restricted generator coverage. Although we include four representative and controllable inpainting/editing pipelines, SA-BENCH does not yet cover a broader range of open-source and commercial generators such as SD3, Recraft, Midjourney, or DALL-E 3, nor does it include dedicated cross-generator evaluation on unseen models. Third, the annotation pipeline uses outlier mitigation to stabilize MOS estimation, which may suppress part of the legitimate subjective variance, and we do not currently report a standard inter-rater coefficient such as Cronbach’s alpha or ICC. Fourth, computational constraints prevented exploring larger model scales (e.g., 32B or 72B parameters) and exhaustive GRPO hyper-parameter optimization, suggesting unreached performance under ideal conditions. Fifth, SA-IQA primarily models perceptual quality from visual evidence. Although dimension-conditioned prompts provide lightweight textual guidance, the current formulation does not fully incorporate user intent, prompt faithfulness, functional requirements, or richer scene semantics.

\paragraph{Social Impact.}
SA-IQA is poised to deliver significant positive societal impact by elevating the quality of AI-generated images (AIGI) and spatial design. It will empower designers to create more aesthetically pleasing and functional spaces, foster greater accessibility of high-quality visual content across various applications, and drive innovation in AI-driven design, ultimately enriching user experiences in both digital and physical realms.

\end{document}